%% file: main.tex
\documentclass{article}
\PassOptionsToPackage{sort}{natbib}
\usepackage{kepler4d,times}

\usepackage[utf8]{inputenc}
\usepackage[T1]{fontenc}
\usepackage[hidelinks]{hyperref}
\hypersetup{
    pdftitle={Kepler4D: Controllable Future Video Generation via 4D Scene State Evolution},
    pdfauthor={Feiran Wang, Bin Duan, Junyi Wu, Gaowen Liu, Yan Yan}
}

\usepackage{url}
\usepackage{booktabs}
\usepackage{tabularx}
\usepackage{array}
\usepackage{amsfonts}
\usepackage{amsmath}
\usepackage{amssymb}
\usepackage{microtype}
\usepackage{xcolor}
\usepackage{graphicx}
\makeatletter
\@ifundefined{Gread@transgrouptrue}{}{\let\Gread@transgrouptrue\Gread@transgroupfalse}
\makeatother
\graphicspath{{evolveworld_nips/figs/images/}}

\definecolor{AutoGreen}{HTML}{4C844A}
\definecolor{EditGold}{HTML}{D69D13}

\title{Kepler4D: Controllable Future Video\\Generation via 4D Scene State Evolution}

\author{
    Feiran Wang\textsuperscript{1},
    Bin Duan\textsuperscript{2},
    Junyi Wu\textsuperscript{1},
    Gaowen Liu\textsuperscript{3},
    Yan Yan\textsuperscript{1}\thanks{\enspace Corresponding author.} \\
    {\normalfont
        \textsuperscript{1}University of Illinois at Chicago,
        \textsuperscript{2}University of Michigan Ann Arbor,
        \textsuperscript{3}Cisco}
}

\begin{document}

\setcounter{footnote}{1}
\maketitle

\input{evolveworld_nips/figs/tex/teaser}
\vspace{-2pt}
\input{evolveworld_nips/sec/0_abstract.tex}
\vspace{-20pt}
\input{evolveworld_nips/sec/1_introduction.tex}
\input{evolveworld_nips/sec/2_relatedwork.tex}
\input{evolveworld_nips/sec/3_method.tex}
\input{evolveworld_nips/sec/4_experiment.tex}

\input{evolveworld_nips/sec/5_conclusion.tex}

\bibliography{references}
\bibliographystyle{kepler4d}

\clearpage
\appendix
\input{evolveworld_nips/sec/appendix.tex}

\end{document}

%% file: evolveworld_nips/figs/tex/teaser.tex
\begin{figure}[!ht]
    \vspace{-2em}
    \centering
    \includegraphics[width=\linewidth, trim={4pt 4pt 4pt 4pt}]{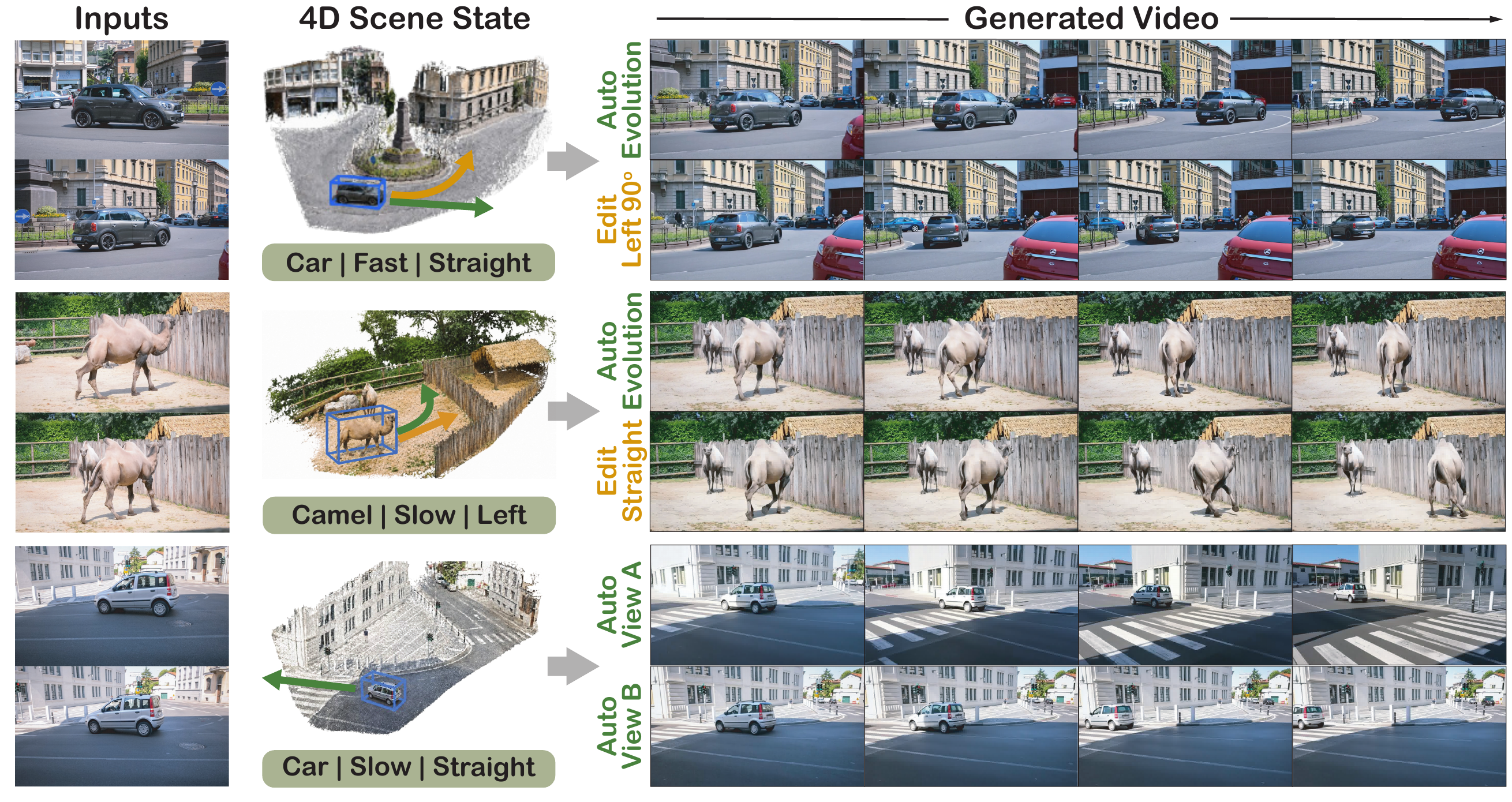}
    \caption{
    Kepler4D enables future video generation through explicit trajectory evolution and control in a 4D scene state.
    Our method organizes static geometry, foreground motion histories, and semantic context into the scene state, predicting scene-aware 3D trajectories to guide generation.
    It supports
    \textcolor{AutoGreen}{auto}matic future evolution,
    language-guided trajectory \textcolor{EditGold}{edit}ing,
    and view-consistent rendering across viewpoints, producing plausible future motion while preserving the reconstructed scene structure.
    }
    \label{fig:teaser}
\end{figure}

%% file: evolveworld_nips/sec/0_abstract.tex
\begin{abstract}
    Video world models aim to preserve scene structure and predict how dynamic objects evolve beyond visual observations. We present Kepler4D, a framework for future video generation through explicit 4D scene state evolution. Given a monocular video, Kepler4D constructs a shared 3D representation of background geometry, object motion histories, coarse spatial supports, and semantic context. Chain-of-Motion summarizes observed motion and uses a vision-language model to select structured speed and heading decisions and decide whether to bound object-center height from below. A deterministic rollout converts these decisions into future object trajectories for inspection and editing before synthesis. We render the evolving proxies into geometric controls for a pretrained video generator, separating coarse object motion from the synthesis of appearance and articulation. Experiments on real-world videos demonstrate that Kepler4D enables controllable object motion and plausible future rollout while preserving scene consistency.
    The project page and code are available at {\urlstyle{same}\url{https://brack-wang.github.io/kepler4d/}}.
\end{abstract}

%% file: evolveworld_nips/sec/1_introduction.tex
\section{Introduction}
\label{sec:intro}
\vspace{-4pt}

Video world models aim to preserve scene structure and predict plausible future dynamics from visual observations~\citep{wang2025evoworld,duan2026liveworld,chen2025deepverse,wu2025video}.
Future video generation is valuable for anticipating scene changes and exploring possible outcomes beyond the observed frames, supporting scenario generation for autonomous driving~\citep{hu2023gaia1,wang2023drivedreamer}, visual planning for robotic manipulation~\citep{zhou2024robodreamer,dharmarajan2026dream2flow}, and action-controllable simulation of interactive environments~\citep{yang2023unisim,bruce2024genie}.

Recent video world models have made progress in maintaining scene structure.
Reconstructed spatial memory, geometric caches, and rendered controls improve consistency across camera viewpoints~\citep{wu2025video,ren2025gen3c,wang2025evoworld}.
Other methods address dynamic subjects and events that evolve outside the current view~\citep{duan2026liveworld,chen2026outofsight}.
Geometry-based controllers also enable camera redirection~\citep{yu2025trajectorycrafter} and joint control of camera and object trajectories~\citep{zheng2026versecrafter}.
However, these scene-consistency and geometric-control approaches typically leave future motion to video synthesis or require external trajectories.
Joint motion prediction and editing therefore benefit from a shared scene representation that connects observed motion and scene context to explicit future object trajectories and their rendering as video generation controls.

We propose \textbf{Kepler4D}, a controllable video world model that observes motion history, predicts scene-aware 3D trajectories, and renders them as motion guidance for future generation.
At its core is a structured 4D scene state that organizes static scene geometry, foreground motion histories, coarse spatial supports, and semantic context in a shared world frame.
Static geometry anchors scene structure, while foreground motion states provide the basis for predicting future object trajectories.
By evolving object centers within this shared 4D state, Kepler4D makes future motion explicit and available for inspection and editing before video synthesis.
This formulation supports automatic future evolution, language-guided trajectory editing, and rendering under different camera viewpoints.

Kepler4D combines state construction, motion reasoning, and controllable rendering.
First, it lifts the input video into background geometry and an object proxy bank, where each proxy stores semantic identity, an observed 3D trajectory, and coarse spatial support.
Second, Chain-of-Motion (CoM) summarizes observed motion and uses a vision-language model to infer structured motion decisions from these summaries and scene context.
Deterministic rollout converts decisions into future 3D trajectories for inspection or editing before synthesis.
Third, predicted or edited trajectories are rendered with background geometry as geometric control maps under a target camera sequence.
These controls guide a pretrained video generator, with background geometry anchoring layout and moving proxies guiding foreground displacement.
The generator synthesizes appearance and local articulation; trajectories expose coarse object motion throughout prediction, editing, and synthesis.

We first evaluate static novel view synthesis on DL3DV, followed by automatic dynamic continuation and language-guided trajectory editing on DAVIS to assess scene rendering and future motion control.
Experiments demonstrate that Kepler4D enables controllable object motion and plausible future rollout while preserving scene consistency.
Ablation studies further show the contributions of observed motion history, semantic context, and rendered trajectory guidance to future video synthesis.

Our contributions are:
\begin{itemize}
    \item We propose Kepler4D, a controllable video world model that integrates future motion prediction and video synthesis through explicit evolution of one shared 4D scene state.
    \item We introduce Chain-of-Motion, which uses observed motion and scene context to infer structured motion decisions and expands them into editable future 3D trajectories through deterministic rollout to guide future video generation with explicit object motion control.
    \item We evaluate automatic evolution, language-guided trajectory editing, and static rendering on real-world videos, ablating motion history, semantic context, and 3D trajectory guidance.
\end{itemize}

%% file: evolveworld_nips/sec/2_relatedwork.tex
\section{Related Work}
\label{sec:relatedwork}

\noindent\textbf{Video World Models and Geometric Control.}
Video world models~\citep{hu2023gaia1,bruce2024genie,chen2025deepverse,wu2025video,wang2025evoworld,duan2026liveworld} aim to simulate scene evolution from visual observations and control signals.
Spatial-memory methods use reconstructed geometry to preserve scene structure across generated views~\citep{wu2025video,ren2025gen3c,wang2025evoworld}.
LiveWorld~\citep{duan2026liveworld} and hybrid-memory approaches~\citep{chen2026outofsight} model dynamic events outside the current view.
For viewpoint control, camera-conditioned generators encode target poses~\citep{he2024cameractrl,wang2024motionctrl}, while TrajectoryCrafter~\citep{yu2025trajectorycrafter} and FreeOrbit4D~\citep{cao2026freeorbit4d} use reconstructed geometry to redirect camera trajectories in monocular videos.
Geometry Forcing improves 3D consistency by aligning intermediate diffusion features with pretrained geometric features during training~\citep{wu2025geometryforcing}.
VerseCrafter supports joint camera and object control by rendering background geometry and per-object 3D Gaussian trajectories into conditioning maps~\citep{zheng2026versecrafter}.
Kepler4D renders editable 3D trajectories predicted from observed motion and scene context as geometric controls for future video generation.

\noindent\textbf{Motion Reasoning and Future Prediction.}
Motion prediction methods~\citep{seff2023motionlm,moon2024visiontrap,yang2025vlipp,yang2025trajvlmgen,zhang2026molmomotion} forecast trajectories from scene observations, motion histories, or language instructions.
MotionLM formulates multi-agent forecasting as autoregressive prediction of discrete motion tokens~\citep{seff2023motionlm}, while VisionTrap incorporates visual features and textual supervision into trajectory prediction~\citep{moon2024visiontrap}.
VLIPP~\citep{yang2025vlipp} and TrajVLM-Gen~\citep{yang2025trajvlmgen} use VLMs to predict coarse motion that guides video synthesis.
SeeU learns continuous 4D dynamics from reconstructed observations to generate videos at unobserved times and viewpoints~\citep{yuan2025seeu}.
MolmoMotion predicts 3D point trajectories from visual history and language goals for downstream control and generation~\citep{zhang2026molmomotion}, while OptiWorld optimizes trajectories for task goals under physical constraints~\citep{yuan2026optiworld}.
Kepler4D combines structured VLM decisions with deterministic 3D rollout to predict scene-aware, editable future object-center trajectories, enabling automatic future evolution and controllable video generation. These trajectories provide direct object motion controls before future video synthesis.

%% file: evolveworld_nips/sec/3_method.tex
\section{Methodology}
\label{sec:method}

Kepler4D reconstructs an observed scene in 3D and predicts future object motion to guide controllable future video generation.
As shown in Fig.~\ref{fig:pipeline}, given an observed video $\mathcal{V}^{\mathrm{src}}=\{I_t\}_{t=0}^{T_o-1}$, we construct a shared 3D representation of static scene geometry, foreground object proxies, and semantic context (\S~\ref{sec:world}, \S~\ref{sec:construct}).
Chain-of-Motion then reasons over motion history and scene context to predict 3D object trajectories for the next $T_f$ frames (\S~\ref{sec:cop}).
Finally, we render the predicted or edited trajectories with static geometry, producing geometric control maps that guide future video synthesis (\S~\ref{sec:render}).

\input{evolveworld_nips/figs/tex/pipeline}

\subsection{4D Scene State and Its Evolution}
\label{sec:world}

Our goal is to construct a proxy world from an observed video and use its predicted evolution to guide future video generation. The representation combines scene geometry for locating and rendering objects with motion histories and semantic context for predicting their motion. We express geometry and motion in a shared reconstructed 3D coordinate frame and define the 4D scene state as
\begin{equation}
    \mathcal{S}=(\mathcal{G},\mathcal{P},\mathcal{A}),
    \qquad
    \mathcal{P}=\{\mathbf{p}_i\}_{i=1}^{N}.
    \label{eq:scene-state}
\end{equation}
Here, $\mathcal{G}$ stores static background geometry, $\mathcal{P}$ contains $N$ foreground object proxies, and $\mathcal{A}$ stores semantic scene context. The background geometry provides a spatial reference for object trajectories and scene rendering. Each proxy $\mathbf{p}_i$ records a semantic label $\ell_i$, an observed 3D center history $\mathcal{H}_i$, and a coarse spatial extent $\mathbf{e}_i$. The center history informs motion prediction, while the spatial extent defines an ellipsoid for rendering the proxy. Scene descriptions and local layout cues in $\mathcal{A}$ provide additional context for predicting object motion. Together, these components provide a common representation for predicting future object motion and rendering the predicted future evolution.

To represent scene changes beyond the observed video, we evolve the proxy world into a sequence of future states by updating object centers along predicted 3D trajectories. Given $T_o$ observed frames and a prediction horizon of $T_f$ frames, the state at each future frame $t=T_o,\ldots,T_o+T_f-1$ is
\begin{equation}
    \widehat{\mathcal{S}}_t
    =
    \left(
    \mathcal{G},
    \{(\ell_i,\widehat{\mathbf{c}}_i^t,\mathbf{e}_i)\}_{i=1}^{N},
    \mathcal{A}
    \right),
    \label{eq:state-evolution}
\end{equation}
where $\widehat{\mathbf{c}}_i^t$ is the predicted center of object $i$ at time $t$. Background geometry, semantic context, and object labels and extents remain fixed throughout the predicted state sequence. Rendering the predicted or edited state sequence under target camera poses converts future scene evolution into geometric controls for video generation. These controls encode scene layout and object displacement, while the video generator synthesizes fine appearance and local articulation over the future frames.

\subsection{Scene State Construction}
\label{sec:construct}

From the observed video, we recover scene geometry and identify foreground objects for motion prediction. A shared 3D coordinate frame describes their motion relative to the scene across frames. Specifically, a pretrained geometry estimator~\citep{lin2025da3} predicts per-frame depth $D_t$, camera intrinsics $K_t$ for image projection, and camera poses $\mathcal{C}_t$ specifying camera position and orientation.

Using this geometric reference, we identify objects and extract scene context for motion prediction. A vision-language model~\citep{bai2025qwen25vltechnicalreport} processes the first and last observed frames to produce object labels, bounding boxes, and a scene description. The labels define $\ell_i$, while the scene description and local layout cues form $\mathcal{A}$. We use the bounding boxes to initialize SAM 2~\citep{ravi2024sam2}, which tracks object regions across frames to produce a mask $\mathcal{M}_i^t$ identifying pixels of object $i$ at time $t$.

To recover object motion in 3D, we combine these masks with the estimated depth and camera parameters to map object pixels into the shared coordinate frame. Denoting this mapping by $\Pi^{-1}$, we estimate each proxy center as the mean of the visible object points reconstructed within its mask:
\begin{equation}
    \mathbf{c}_i^t
    =
    \frac{1}{|\mathcal{M}_i^t|}
    \sum_{u\in\mathcal{M}_i^t}
    \Pi^{-1}\!\left(u,D_t(u),K_t,\mathcal{C}_t\right).
    \label{eq:lifting}
\end{equation}
Here, $u$ denotes a pixel location and $|\mathcal{M}_i^t|$ is the number of pixels in the object mask. These observed centers form the motion history $\mathcal{H}_i$ for future trajectory prediction. We estimate the spatial extent $\mathbf{e}_i$ from reconstructed object points to define the ellipsoid rendered at each predicted future position.

Finally, we construct static background geometry from multiple input frames to retain the observed surroundings during future object motion. Using the same mapping, we transform pixels outside the foreground masks into the shared 3D coordinate frame, and aggregate the resulting points into $\mathcal{G}$. This aggregation incorporates background regions visible in different frames into the scene used for future rendering. Together, these steps construct $\mathcal{S}$, providing motion histories and semantic context for subsequent trajectory prediction and scene geometry for rendering predicted scene evolution.

\subsection{Chain-of-Motion Inference}
\label{sec:cop}

To evolve the reconstructed scene, we use motion histories to describe recent movement and scene context to infer how it may continue or change. Inspired by structured spatial reasoning~\citep{zhen2026layoutr1}, we introduce Chain-of-Motion (CoM), which first extracts motion attributes, then infers motion decisions, and finally computes future 3D trajectories. These decisions translate semantic reasoning about future behavior into numerical position updates for rendering the future 3D scene.

\noindent\textbf{Motion-state extraction.}
To support motion reasoning, we encode each object's recent movement and local scene conditions in a compact motion state $\mathbf{m}_i$. The function $\Psi_{\mathrm{state}}$ extracts these attributes from the object proxy $\mathbf{p}_i$ and semantic context $\mathcal{A}$ for subsequent object motion prediction:
\begin{equation}
    \mathbf{m}_i
    =
    \Psi_{\mathrm{state}}(\mathbf{p}_i,\mathcal{A})
    =
    \left[
    \mathbf{c}_i^{T_o-1},
    \mathbf{v}_i,
    s_i,
    \mathbf{a}_i,
    \kappa_i,
    h_i,
    \mathbf{e}_i,
    q_i
    \right],
    \label{eq:motion-state}
\end{equation}
where $\mathbf{c}_i^{T_o-1}$ is the latest observed center, while $(\mathbf{v}_i,s_i,\mathbf{a}_i,\kappa_i)$ denote velocity, speed, acceleration, and turning trend estimated from $\mathcal{H}_i$. The remaining entries are the height threshold $h_i$ used by the optional lower-bound constraint, object extent $\mathbf{e}_i$, and semantic cues $q_i$ about the category or surroundings. These attributes relate recent motion to the scene for subsequent motion decisions.

\noindent\textbf{Structured motion decisions.}
Given $\mathbf{m}_i$, a pretrained VLM interprets recent movement in its scene context to infer future behavior. We express this behavior as structured decisions about object speed, motion direction, and whether to apply a lower height bound. With an optional language instruction $\ell$, the inference module $\Phi_{\mathrm{CoM}}$ produces these decisions together with a textual motion description:
\begin{equation}
    (\alpha_i,\delta_i,\eta_i,d_i)
    =
    \Phi_{\mathrm{CoM}}(\mathbf{m}_i,\mathcal{A},\ell),
    \label{eq:com-decision}
\end{equation}
where $\alpha_i$ specifies whether to maintain, increase, or decrease speed, while $\delta_i$ selects whether to continue in the current direction or turn. The decision $\eta_i$ determines whether to apply a lower bound to the predicted object-center height during trajectory computation. The motion description $d_i$ guides video synthesis, whereas structured decisions govern numerical position updates. These decisions can also be edited to specify alternative future object motion within the same reconstructed scene.

\noindent\textbf{Deterministic trajectory rollout.}
To render the evolving scene, we convert these decisions into a 3D object position for each future frame. Deterministic trajectory rollout applies fixed numerical update rules from the latest observed position and velocity. Specifically, $\Gamma_{\mathrm{motion}}$ maps the motion decisions to numerical parameters $\mathbf{u}_i$, which $\mathcal{I}_{\mathrm{traj}}$ uses to compute positions over the next $T_f$ frames:
\begin{equation}
    \mathbf{u}_i
    =
    \Gamma_{\mathrm{motion}}
    \left(
    \alpha_i,\delta_i,\eta_i,h_i
    \right),
    \qquad
    \widehat{\mathcal{T}}_i
    =
    \mathcal{I}_{\mathrm{traj}}
    \left(
    \mathbf{c}_i^{T_o-1},
    \mathbf{v}_i,
    \mathbf{u}_i;
    T_f
    \right),
    \label{eq:traj-rollout}
\end{equation}
At each step, speed and heading decisions adjust the displacement used to update the object center. When enabled, the height constraint imposes $h_i$ as a lower bound on the predicted object-center height. We then apply temporal smoothing to reduce abrupt changes between successive predicted positions. The resulting trajectory $\widehat{\mathcal{T}}_i=\{\widehat{\mathbf{c}}_i^t\}_{t=T_o}^{T_o+T_f-1}$ supplies the centers in Eq.~\ref{eq:state-evolution}, converting predicted or edited motion into future scene configurations for rendering under target camera poses.

\subsection{Future Video Generation}
\label{sec:render}

We generate future video by rendering the evolving proxy world under target camera poses and using the resulting controls to guide video synthesis.
The future scene states specify scene layout and object motion, while the video generator synthesizes fine appearance details absent from the coarse proxy representation.
We adopt the geometric control representation and pretrained video generation components of VerseCrafter~\citep{zheng2026versecrafter} to guide future synthesis from the scene states.

For each future state, we render complementary background and foreground controls under the corresponding target camera pose.
The background controls encode the appearance and depth of the static scene.
Foreground objects are represented by coarse 3D Gaussian ellipsoids whose centers follow the predicted or edited trajectories and whose sizes are determined by the proxy supports.
Rendering these proxies produces foreground color, depth, and motion masks that indicate where objects should appear in each future frame.
Both sets of controls are derived from the shared world frame, allowing camera motion and object motion to jointly guide synthesis across all future frames.
Here, changing the target camera pose changes the projection of the reconstructed state, whereas editing an object trajectory changes its center positions within that state.
The same explicit motion plan can therefore be rendered under alternative target camera paths, and each object-motion edit remains defined in the shared scene rather than in the image coordinates of a particular camera view.

To preserve visual continuity, we use the last observed frame $I_{T_o-1}$ as an appearance reference for the scene and its foreground objects.
We also construct a text prompt $p_{\mathrm{text}}$ from CoM's motion descriptions $d_i$ to complement the geometric controls with semantic guidance for object motion.
The generator combines these appearance, geometric, and motion cues to generate the future video:
\begin{equation}
    \widehat{\mathcal{V}}
    =
    \Phi_{\mathrm{VG}}\!\left(
    I_{T_o-1},
    \mathcal{R}\!\left(
    \{\widehat{\mathcal{S}}_t\}_{t=T_o}^{T_o+T_f-1},
    \mathcal{C}^{\mathrm{tar}}
    \right),
    p_{\mathrm{text}}
    \right).
    \label{eq:render-video}
\end{equation}
Here, $\mathcal{R}$ renders the background and foreground controls for each state under the target camera sequence $\mathcal{C}^{\mathrm{tar}}$, and $\Phi_{\mathrm{VG}}$ is the pretrained generator.
The implementation appendix summarizes the model components and inference settings used here.
By evolving the proxy world, Kepler4D supports automatic video continuation, while trajectory edits allow users to specify alternative object motion.

%% file: evolveworld_nips/figs/tex/pipeline.tex
\begin{figure}[t]
    \centering
    \includegraphics[width=1.0\textwidth]{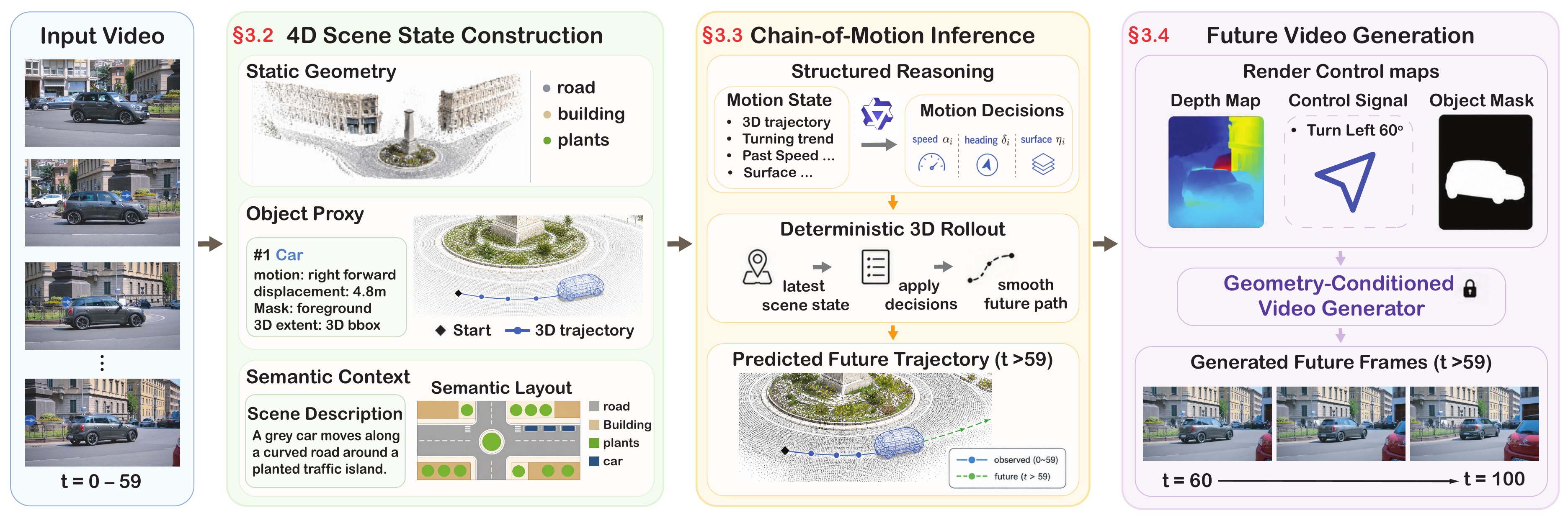}
    \caption{
    \textbf{Overview of Kepler4D.}
    Given an observed video, Kepler4D constructs a structured 4D scene state comprising static scene geometry, foreground motion histories, and semantic context.
    Chain-of-Motion reasons over the observed motion and scene context to infer structured motion decisions, which a deterministic 3D rollout expands into future object trajectories.
    These trajectories are rendered with static geometry into depth maps, object control signals, and masks to guide a pretrained video generator.
    This design separates explicit object motion prediction from video synthesis, enabling controllable generation of future videos while preserving the scene structure.
    }
    \label{fig:pipeline}
    \vspace{-0.5em}
\end{figure}

%% file: evolveworld_nips/sec/4_experiment.tex
\section{Experiments}
\label{sec:experiment}

We first introduce the datasets and evaluation metrics.
We then evaluate Kepler4D across three settings: static view synthesis in \S~\ref{sec:static_nvs}, automatic future video generation in \S~\ref{sec:unconditioned_dynamic}, and controllable future video generation in \S~\ref{sec:controllable_dynamic}.
These experiments assess scene rendering under camera motion, continuation of observed object dynamics, and control over future object movement, respectively.

\noindent\textbf{Datasets.}
We evaluate static view synthesis on the 55-scene DL3DV-Evaluation release~\citep{ling2024dl3dv}, using the first 49 frames as observations and the provided camera trajectories for future viewpoints.
Dynamic continuation and motion editing draw on the 90 sequences of the DAVIS 2017 train and validation splits~\citep{perazzi2016benchmark,ponttuset2017davis}.
For automatic continuation, we split 38 selected sequences from this pool into observed prefixes and held-out continuations.
We estimate camera trajectories from the full sequences with Depth Anything 3~\citep{lin2025da3} and use future poses as external rendering conditions.
Within each setting, all methods share observed frames; camera-controllable methods share the target camera path.
Appendix~\ref{app:dataset_usage} describes the dataset usage and temporal alignment procedures for these evaluations and the role of the target camera paths.

\noindent\textbf{Evaluation Metrics.}
We evaluate reference similarity, visual quality, and motion accuracy.
PSNR, SSIM~\citep{wang2004ssim}, and LPIPS~\citep{zhang2018perceptual} measure similarity to held-out frames.
FID~\citep{heusel2017gans} compares feature distributions of generated and reference frames.
VBench~\citep{huang2024vbench} assesses visual quality, temporal consistency, and motion activity.
Dynamic Degree measures activity, not accuracy.
For dynamic continuation, Flow-EPE measures the full-image optical-flow difference between generated and reference videos, reflecting camera and object motion, while RotErr measures camera rotation error.
For motion editing, we report the camera-yaw-based TurnErr proxy and VBench Imaging Quality.
TurnErr reflects camera turning, not object heading independent of camera motion.
PSNR measures similarity to the recorded continuation, which may not reflect the requested motion.
Appendix~\ref{app:metrics} provides the metric definitions and evaluation protocols used here and explains the interpretation of each of the reported scores.

\input{evolveworld_nips/sec/4_static}

\input{evolveworld_nips/sec/4_automatic}

\input{evolveworld_nips/sec/4_control}

\subsection{Ablation Study}
\label{sec:analysis}

\noindent\textbf{Trajectory perturbation study.}
We evaluate whether coarse trajectory-centered control can steer
trajectory-conditioned rendering by perturbing the ground-truth future 3D
trajectory with the input video, target camera path, and generator fixed.
As shown in Table~\ref{tab:traj_perturb}, GT trajectory control yields the lowest
AngleErr and SurfErr and the highest imaging quality and dynamic degree.
AngleErr increases from 5.78 to 11.37 when rotating the trajectory $60^\circ$ left
and to 26.92 under the downward perturbation, indicating reduced adherence to
the specified control directions. SurfErr also increases from 0.040 to 0.048
and 0.056, respectively, while imaging quality remains below that of the
unperturbed control. These results characterize how directional adherence and
synthesis quality change under trajectory perturbations, with the downward
perturbation producing the largest reported directional error.

\noindent\textbf{Scene state component ablation.}
We ablate components of the 4D scene state while keeping the same input
video, target camera path, and rendering pipeline. As shown in
Table~\ref{tab:proxy_ablation}, removing motion history causes the largest
increase in Flow-EPE and drop in DynDeg. Removing semantic context also
increases Flow-EPE. Removing surface cues has a smaller effect on Flow-EPE,
but increases SurfErr and reduces imaging quality. Within the evaluated generation
pipeline, these comparisons support using motion history, semantic context,
and surface cues together to guide future generation.

\input{evolveworld_nips/tab/controltable}
\input{evolveworld_nips/tab/ablation}

%% file: evolveworld_nips/sec/4_static.tex
\input{evolveworld_nips/figs/tex/staticCompare}
\subsection{Static Novel View Synthesis}
\label{sec:static_nvs}

We first evaluate the static setting on the DL3DV dataset, where foreground evolution is absent and
the task reduces to rendering a static scene under future camera viewpoints.
When no dynamic foreground object is detected, Kepler4D uses an empty object
proxy bank and relies solely on the reconstructed background state for
geometry-conditioned video synthesis. This setting evaluates background
reconstruction and camera-conditioned generation, without testing CoM motion reasoning itself.

\noindent\textbf{Results.}
As shown in Figure~\ref{fig:static_compare}, in indoor
and outdoor scenes, Kepler4D better preserves scene layout and fine
textures under target camera motion for wide-baseline
views, where baselines show geometric distortion, surface
inconsistency, or texture drift. These results support our
static background representation and geometry-conditioned rendering for static novel view synthesis.

Table~\ref{tab:dl3dv_static} shows competitive overall static generation results for
Kepler4D. Reconstruction gains indicate that scene-state rendering better
preserves scene structure in the target views, while lower LPIPS and FID suggest that the generated
views retain finer appearance details and visual realism.

Although
GEN3C achieves the highest Motion Smoothness, Kepler4D performs better on most
reconstruction and VBench metrics, balancing
viewpoint accuracy, visual quality, and scene consistency under target camera
motion. Efficiency results indicate a runtime--memory trade-off under the
evaluated configurations. Kepler4D uses 62.46 GB of peak GPU memory, compared
with 37.81--54.73 GB for the baselines, while also recording the lowest reported
per-clip inference time of 5.2 minutes.

\input{evolveworld_nips/tab/staticCompare}

%% file: evolveworld_nips/figs/tex/staticCompare.tex
\begin{figure}[t]
    \centering
    \includegraphics[width=0.94\linewidth]{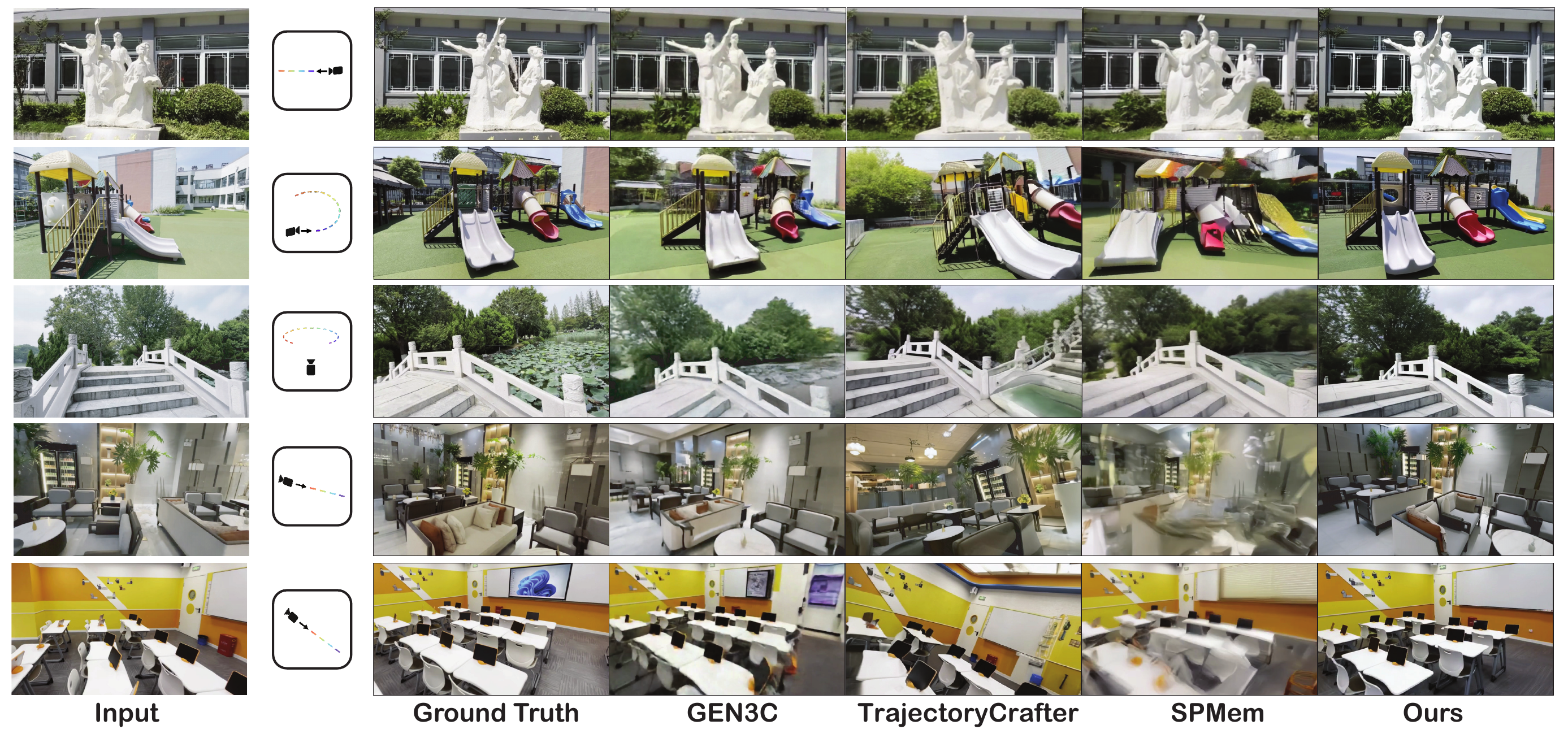}
    \caption{\textbf{Static novel view synthesis on DL3DV.}
    We compare Kepler4D with GEN3C, TrajectoryCrafter, and SPMem on indoor and outdoor scenes under target camera motion.
    Each row shows an input view, the target camera path, the held-out reference view, and the generated results.
    In these examples, Kepler4D preserves the scene layout and appearance details across viewpoint changes.}
    \label{fig:static_compare}
    \vspace{-0.5em}
\end{figure}

%% file: evolveworld_nips/tab/staticCompare.tex
\begin{table}[t]
    \vspace{-1.0em}
    \centering
    \footnotesize
    \renewcommand{\arraystretch}{1.}
    \setlength{\tabcolsep}{2.5pt}
    \caption{
    \textbf{Static view synthesis on DL3DV~\citep{ling2024dl3dv}.}
    We compare Kepler4D with three baselines using reference fidelity,
    FID, and VBench~\citep{huang2024vbench} metrics.
    We also report peak GPU memory and per-clip inference time
    under the evaluated model and inference configurations.
    }
    \resizebox{1.0\textwidth}{!}{
    \begin{tabular}{@{}lccc|ccccc|cc@{}}
        \toprule
        & \multicolumn{3}{c|}{Reconstruction Fidelity} & \multicolumn{5}{c|}{Generation Quality} & \multicolumn{2}{c}{Efficiency} \\
        \cmidrule(lr){2-4} \cmidrule(lr){5-9} \cmidrule(l){10-11}
        \textbf{Method}
        & {PSNR $\uparrow$} & {SSIM $\uparrow$} & {LPIPS $\downarrow$}
        & {FID $\downarrow$} & {Imag. $\uparrow$} & {MotSm. $\uparrow$} & {BgC. $\uparrow$} & {Overall $\uparrow$}
        & {Mem. (GB)} & {Time (min)} \\
        \midrule
        GEN3C~\cite{ren2025gen3c}                  & 11.67 & 0.2963 & 0.5706 & 63.99 & \underline{0.6991} & \textbf{0.9863} & 0.9314 & 0.8041 & 54.73 & 43  \\
        TrajectoryCrafter~\cite{yu2025trajectorycrafter} & 11.90 & 0.3114 & 0.5791 & 74.23 & 0.6709 & 0.9712 & \underline{0.9372} & 0.8005 & 37.81 & 10.4 \\
        SPMem~\cite{wu2025video}                   & \underline{13.19} & \underline{0.3611} & \underline{0.5273} & \underline{56.92} & 0.5819 & \underline{0.9776} & 0.9134 & \underline{0.8175} & 45.57 & 9.7 \\
        \textbf{Ours}                              & \textbf{13.55} & \textbf{0.3675} & \textbf{0.3995} & \textbf{41.37} & \textbf{0.7606} & 0.9745 & \textbf{0.9448} & \textbf{0.8322} & 62.46 & 5.2 \\
        \bottomrule
    \end{tabular}
    }
    \label{tab:dl3dv_static}
    \vspace{-8pt}
\end{table}

%% file: evolveworld_nips/sec/4_automatic.tex
\subsection{Automatic Dynamic Evolution}
\label{sec:unconditioned_dynamic}

We next evaluate dynamic world evolution, where the task is to continue
foreground motion from a monocular video while rendering the scene along a
target future camera path. Experiments are conducted on DAVIS, where all
methods receive the same observed video and target camera trajectory, but no
object trajectories, mask sequences, or language motion commands. This setting
tests whether object evolution can be inferred from visual
observations without externally specified motion control.

\noindent\textbf{Results.}
Table~\ref{tab:davis_dynamic} evaluates Kepler4D in a challenging setting where
foreground motion is inferred from the observed video without object
trajectories or language motion commands. Kepler4D obtains favorable
motion-fidelity scores while maintaining competitive reconstruction and
generation quality. The lower Flow-EPE indicates closer overall image motion
to the reference video, while the lower RotErr indicates more accurate following
of the target camera rotation. The reconstruction and perceptual scores show
that the complete pipeline maintains competitive visual quality during automatic
continuation. SPMem maintains competitive pixel-level fidelity but has higher
image-motion and camera-rotation errors in this evaluation. VerseCrafter achieves
the lowest LPIPS, while its Flow-EPE and RotErr are higher than those of Kepler4D.
Together, these results support explicit trajectory guidance within the complete
pipeline for automatic future generation from video inputs.

Figure~\ref{fig:dynamic_compare} provides a qualitative comparison of these
differences. The observed video already shows a parkour runner vaulting over a low wall. The future scene contains a
railing along the runner's route, and another vault is a plausible continuation
given the observed activity and layout. The reference video and
Kepler4D both show the person moving forward and vaulting over the railing.
TrajectoryCrafter shows a viewpoint change without a vault, SPMem shows apparent
intersections between the person and railing, and VerseCrafter shows a lateral
deviation before the vault. This example illustrates a plausible continuation
produced by the full pipeline: predicted center trajectories provide coarse
motion guidance, while the video generator synthesizes appearance and local
articulation.

\input{evolveworld_nips/figs/tex/dynamicCompare}

\input{evolveworld_nips/tab/dynamicCompare}

%% file: evolveworld_nips/figs/tex/dynamicCompare.tex
\begin{figure}[t]
    \centering
    \includegraphics[width=1.0\linewidth]{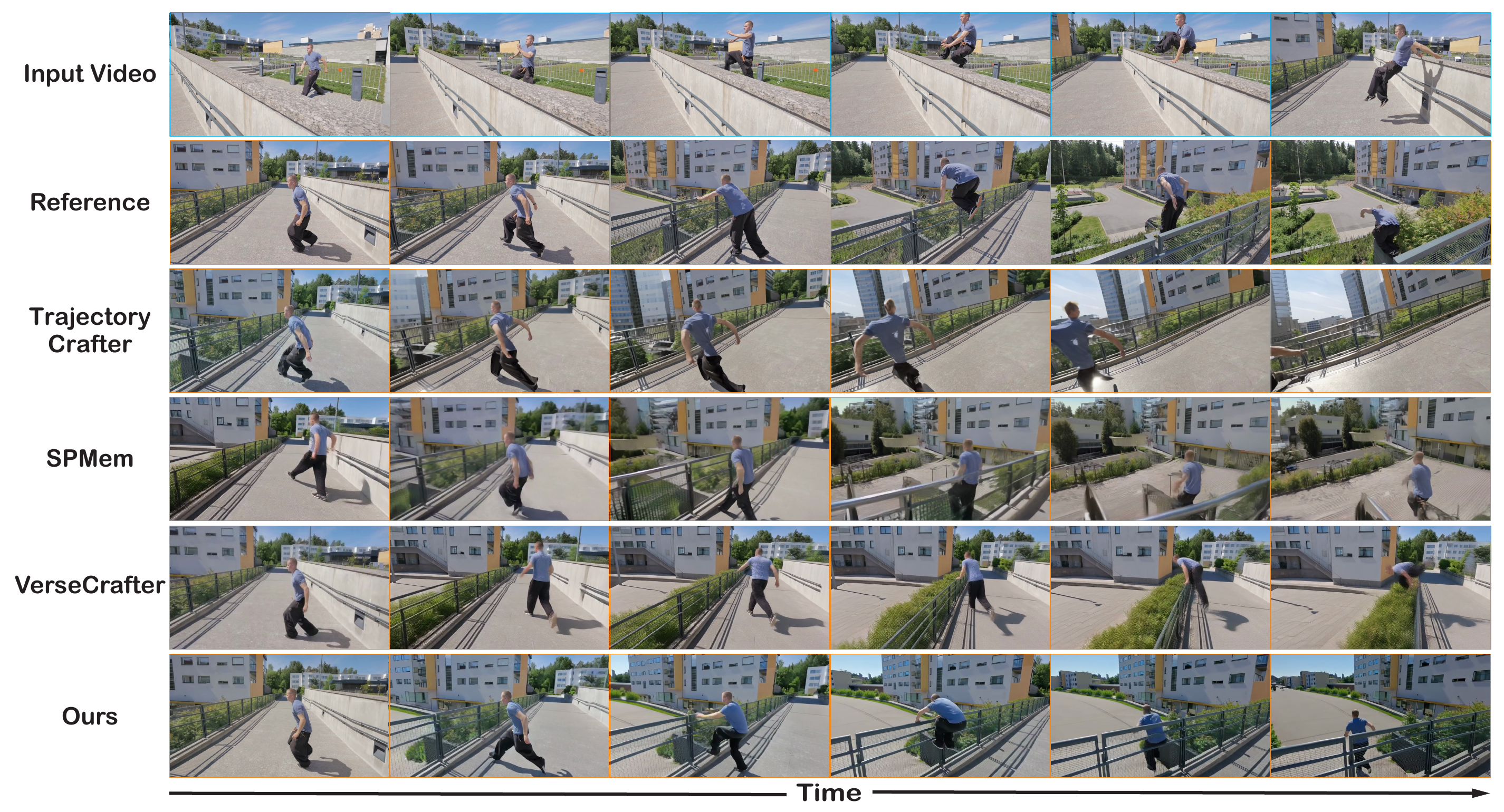}
    \vspace{-2.0em}
    \caption{
    \textbf{Automatic future evolution.}
    All methods use the input video without externally supplied object trajectories or motion commands.
    Methods with camera control are given the same target camera path.
    Kepler4D depicts the runner moving forward and vaulting over the railing, as in the reference.
    TrajectoryCrafter shows a viewpoint change without a vault.
    SPMem shows apparent intersections between the runner and railing.
    VerseCrafter shows a lateral deviation before the vault.
    }
    \label{fig:dynamic_compare}
    \vspace{-1.0em}
\end{figure}

%% file: evolveworld_nips/tab/dynamicCompare.tex
\begin{table}[t]
    \vspace{-0.5em}
    \centering
    \footnotesize
    \renewcommand{\arraystretch}{1.}
    \setlength{\tabcolsep}{2.5pt}
    \caption{
    \textbf{Automatic evolution on DAVIS~\citep{ponttuset2017davis}.}
    All methods receive the same observed video and camera path. We report reconstruction fidelity, generation, and motion
    fidelity.
    }
    \resizebox{1.0\textwidth}{!}{
    \begin{tabular}{@{}lccc|cccc|cc@{}}
        \toprule
        & \multicolumn{3}{c|}{Reconstruction Fidelity} & \multicolumn{4}{c|}{Generation Quality} & \multicolumn{2}{c}{Motion Fidelity} \\
        \cmidrule(lr){2-4} \cmidrule(lr){5-8} \cmidrule(lr){9-10}
        \textbf{Method}
        & {PSNR $\uparrow$} & {SSIM $\uparrow$} & {LPIPS $\downarrow$}
        & {FID $\downarrow$} & {Imag. $\uparrow$} & {DynDeg $\uparrow$} & {Overall $\uparrow$}
        & {Flow-EPE $\downarrow$} & {RotErr $\downarrow$} \\
        \midrule
        TrajectoryCrafter~\cite{yu2025trajectorycrafter}
        & 14.349 & 0.3593 & 0.5145
        & \underline{214.13} & 0.6595 & 0.1622 & 0.7064
        & 17.409 & \underline{7.5400} \\
        SPMem~\cite{wu2025video}
        & \underline{14.641} & \underline{0.3837} & 0.5102
        & 265.13 & 0.5870 & \textbf{0.7838} & \underline{0.7832}
        & 20.138 & 18.549 \\
        VerseCrafter~\cite{zheng2026versecrafter}
        & 14.090 & 0.3649 & \textbf{0.4556}
        & 303.80 & \underline{0.6896} & 0.6486 & 0.7773
        & \underline{15.400} & 8.9610 \\
        \textbf{Ours}
        & \textbf{14.921} & \textbf{0.3924} & \underline{0.4561}
        & \textbf{202.64} & \textbf{0.6970} & \textbf{0.7838} & \textbf{0.8029}
        & \textbf{15.220} & \textbf{5.0200} \\
        \bottomrule
    \end{tabular}
    }
    \label{tab:davis_dynamic}
    \vspace{-1.5em}
\end{table}

%% file: evolveworld_nips/sec/4_control.tex
\input{evolveworld_nips/figs/tex/controlGenerate}
\subsection{Controllable Dynamic Generation}
\label{sec:controllable_dynamic}

We further evaluate language-guided dynamic control on DAVIS. Unlike
\S~\ref{sec:unconditioned_dynamic}, where motion is inferred without user
guidance, this setting provides high-level commands such as going straight or
turning left by a desired angle. Kepler4D converts these commands into
object-level motion plans and renders the edited scene state using
geometry-conditioned control maps. Baselines receive the same language commands,
but not Kepler4D's generated trajectory visualizations or control maps.

\noindent\textbf{Results.}
Figure~\ref{fig:control_compare} illustrates distinct future generations from the
same input video under different language commands. Kepler4D translates these
commands into object-level trajectories and renders them as spatial guidance,
producing command-dependent changes in the generated motion. Wan Inpaint~\citep{wan2025wan} and
SPMem receive the same commands without these rendered trajectory controls and
exhibit different motion responses in the displayed examples.
Table~\ref{tab:control_compare} reports the highest PSNR and imaging quality for
Kepler4D, together with the lowest TurnErr under the stated evaluation protocol. These scores do not directly measure future object-heading accuracy.

\noindent\textbf{Effect of motion conditioning.}
We examine what motion descriptions and rendered trajectories contribute to generation.
The Qwen Prompt baseline uses the scene description from scene-state construction, which describes the observed objects within their surroundings.
CoM w/o traj.\ instead provides a motion description that tells the generator how those objects are expected to move.
The CoM w/ traj.\ variant in this comparison additionally renders the predicted trajectories, specifying where the objects should appear within the scene over future frames.
This gives the generator spatial guidance to accompany the motion description.
Table~\ref{tab:cop_ablation} shows that CoM's motion description improves FID and imaging quality over scene prompting, with further gains after adding rendered trajectories.
These results suggest that describing future movement provides guidance for synthesis, while specifying its spatial progression through rendered trajectories contributes beyond text alone.

%% file: evolveworld_nips/figs/tex/controlGenerate.tex
\begin{figure}[htpb]
    \centering
    \includegraphics[width=1.0\linewidth]{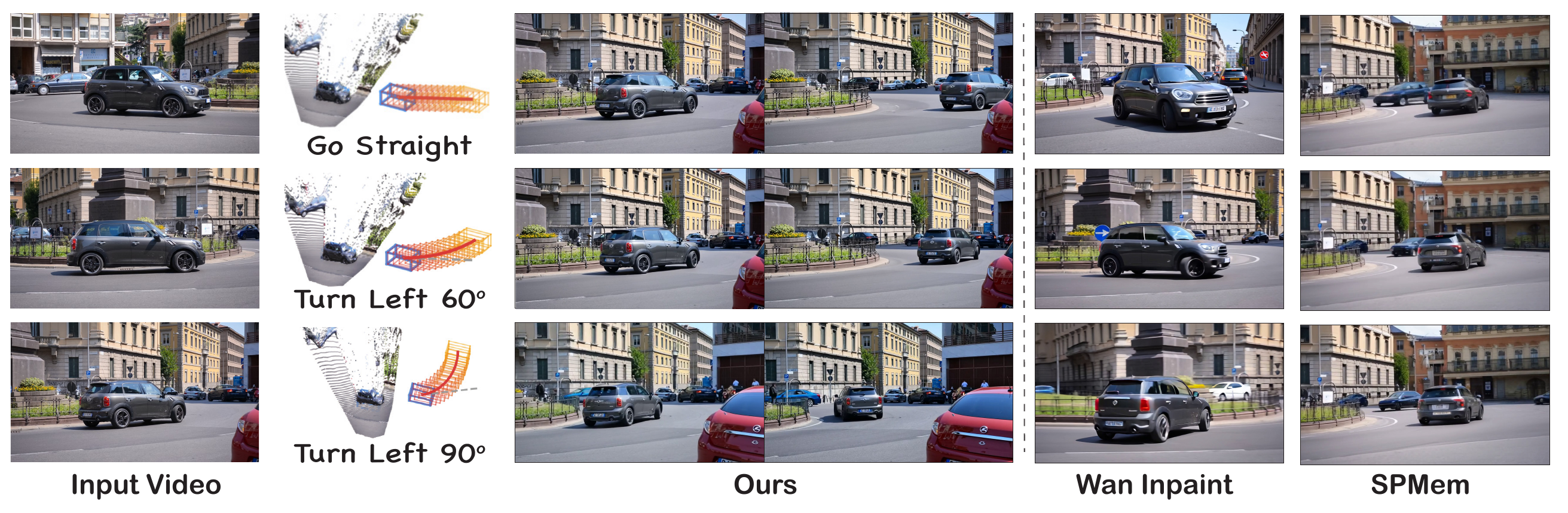}
    \vspace{-2.0em}

    \caption{
    \textbf{Language-guided controllable future generation.}
    Given the same input video, the user specifies the intended motion through a high-level language command.
    Kepler4D infers an object-level trajectory from the command, shown in the second
    column. Baselines receive the same
    language command but not the inferred trajectory or control maps. With explicit
    trajectory control, Kepler4D better realizes the intended object motion while
    preserving scene layout and  consistency.
    }
    \label{fig:control_compare}
    \vspace{-1.0em}
\end{figure}

%% file: evolveworld_nips/tab/controltable.tex
\begin{table}[t]
    \vspace{-2.0em}
    \centering
    \footnotesize
    \renewcommand{\arraystretch}{1.0}
    \setlength{\tabcolsep}{3.5pt}

    \begin{minipage}{0.48\linewidth}
        \centering
        \caption{
        \textbf{Language-guided dynamic control under identical inputs.}
        All methods receive the same observed video and language command.
        }
        \resizebox{1.0\linewidth}{!}{
        \begin{tabular}{@{}lcccc@{}}
            \toprule
            \textbf{Method} & PSNR$\uparrow$ & TurnErr$\downarrow$ & Imag.$\uparrow$ & DynDeg$\uparrow$ \\
            \midrule
            Wan Inpaint & 11.52 & 278.52 & 0.7177 & \textbf{0.831} \\
            SPMem       & 15.18 & 127.49 & 0.6855 & 0.774 \\
            \textbf{Ours} & \textbf{15.27} & \textbf{85.720} & \textbf{0.7970} & 0.792 \\
            \bottomrule
        \end{tabular}
        }
        \label{tab:control_compare}
    \end{minipage}
    \hfill
    \begin{minipage}{0.48\linewidth}
        \centering
        \caption{
        \textbf{Ablation on trajectory-conditioned control.}
        We compare language prompting, CoM reasoning without and with trajectory
        rendering.
        }
        \resizebox{0.88\linewidth}{!}{
        \begin{tabular}{@{}lcccc@{}}
            \toprule
            \textbf{Variant}
            & SSIM$\uparrow$ & PSNR$\uparrow$ & FID$\downarrow$ & Imag.$\uparrow$ \\
            \midrule
            Qwen prompt & 0.2622 & 15.175 & 79.53 & 0.747 \\
            CoM w/o traj. & 0.2598 & 15.106 & 60.49 & 0.761 \\
            \textbf{CoM w/ traj.} & \textbf{0.2692} & \textbf{15.189} & \textbf{48.76} & \textbf{0.765} \\
            \bottomrule
        \end{tabular}
        }
        \label{tab:cop_ablation}
    \end{minipage}
    \vspace{-1.0em}
\end{table}

%% file: evolveworld_nips/tab/ablation.tex
\begin{table}[t]
    \centering
    \footnotesize
    \renewcommand{\arraystretch}{1.0}
    \setlength{\tabcolsep}{3.5pt}

    \begin{minipage}{0.48\linewidth}
        \centering
        \caption{
        \textbf{Trajectory perturbation study.}
        We perturb the GT trajectory to evaluate the robustness of trajectory-conditioned video rendering.
        }
        \resizebox{1.0\linewidth}{!}{
        \begin{tabular}{@{}lcccc@{}}
            \toprule
            \textbf{Control} & AngleErr$\downarrow$ & SurfErr$\downarrow$ & Imag.$\uparrow$ & DynDeg$\uparrow$ \\
            \midrule
            GT traj.          & \textbf{5.78}  & \textbf{0.040} & \textbf{0.6939} & \textbf{0.7862} \\
            Left $60^\circ$  & 11.37 & 0.048 & 0.6815 & 0.7562 \\
            Down $60^\circ$  & 26.92 & 0.056 & 0.6848 & 0.7116 \\
            \bottomrule
        \end{tabular}
        }
        \label{tab:traj_perturb}
    \end{minipage}
    \hfill
    \begin{minipage}{0.48\linewidth}
        \centering
        \caption{
        \textbf{Scene state component ablation.}
        We remove each scene-state component to assess its role in motion reasoning and video synthesis.
        }
        \resizebox{1.0\linewidth}{!}{
        \begin{tabular}{@{}lcccc@{}}
            \toprule
            \textbf{Variant}
            & Flow-EPE$\downarrow$ & SurfErr$\downarrow$ & DynDeg$\uparrow$ & Imag.$\uparrow$ \\
            \midrule
            w/o Motion Hist. & 18.742 & 0.049 & 0.6421 & 0.6954 \\
            w/o Sem. Context & 16.831 & 0.057 & 0.7616 & 0.6918 \\
            w/o Surface Cue  & 15.364 & 0.054 & 0.7593 & 0.6872 \\
            \textbf{Full}    & \textbf{15.220} & \textbf{0.043} & \textbf{0.7838} & \textbf{0.6970} \\
            \bottomrule
        \end{tabular}
        }
        \label{tab:proxy_ablation}
    \end{minipage}
    \vspace{-1.0em}
\end{table}

%% file: evolveworld_nips/sec/5_conclusion.tex
\section{Conclusion}
\label{sec:conclusion}

Kepler4D presents a structured approach to dynamic video world modeling by
representing future evolution as explicit object-centric 3D trajectories inferred
from monocular observations. Its 4D scene state separates scene structure,
foreground motion, and semantic context into a shared, editable intermediate
representation, while Chain-of-Motion reasoning and deterministic rollout expose
future dynamics as interpretable control signals before rendering. By combining
trajectory-grounded reasoning with geometry-conditioned video generation,
Kepler4D supports controllable future object motion under target camera paths,
while the video generator supplies appearance details and local articulation.
Experiments across static rendering, automatic dynamic evolution, and language-guided
control demonstrate the utility of an explicit motion representation for future video generation.

\section*{AI Use Statement}
\addcontentsline{toc}{section}{AI Use Statement}

Generative AI assistants were used only to polish the manuscript for clarity, grammar, and presentation. The research ideas, methods, experiments, and conclusions were developed and verified by the authors. Pretrained vision-language models used as components of our method are explicitly described in the corresponding methodology and implementation sections. All AI-assisted manuscript edits were reviewed by the authors, who take responsibility for the final content of this work.

\section*{Ethics Statement}
\addcontentsline{toc}{section}{Ethics Statement}

Our experiments use existing public research datasets and do not involve new human-subject data collection. Kepler4D is developed for research on controllable future video generation and scene evolution. As generated videos represent possible continuations rather than verified future events, they should be interpreted accordingly. Applications in safety-critical or real-world decision-making settings would require additional validation and appropriate safeguards.

\section*{Reproducibility Statement}
\addcontentsline{toc}{section}{Reproducibility Statement}

We provide implementation and evaluation details in the appendix, including model configurations, scene-state construction, trajectory rollout, geometric controls, evaluation protocols, and metric definitions. We will release the code and evaluation to support reproduction and further research.

%% file: evolveworld_nips/sec/appendix.tex
\setcounter{equation}{0}
\setcounter{table}{0}
\renewcommand{\theequation}{A\arabic{equation}}
\renewcommand{\thetable}{A\arabic{table}}
\renewcommand{\theHequation}{appendix.\arabic{equation}}
\renewcommand{\theHtable}{appendix.\arabic{table}}

\section*{Technical Appendices and Supplementary Material}
\addcontentsline{toc}{section}{Technical Appendices and Supplementary Material}

This appendix provides additional details on the components, motion interface,
geometric controls, and evaluation of Kepler4D. We retain the notation of the
main paper: $\mathcal{S}=(\mathcal{G},\mathcal{P},\mathcal{A})$ denotes the 4D scene state,
$\mathcal{H}_i$ is the observed center history of object $i$, $\mathbf{e}_i$ is
its coarse spatial extent, and $\widehat{\mathcal{T}}_i$ denotes its future
center trajectory. Observed frames are indexed by
$t=0,\ldots,T_o-1$, and the future horizon contains $T_f$ frames.

\subsection*{Appendix Contents}
\begingroup
\hypersetup{hidelinks}
\setlength{\parindent}{0pt}
\setlength{\parskip}{3pt}
\newcommand{\appcontentsline}[3]{\noindent\hspace*{#1}{#2\hyperref[#3]{\makebox[2.6em][l]{\ref*{#3}}\nameref*{#3}}\nobreak\dotfill\nobreak\hyperref[#3]{\pageref*{#3}}\par}}
\appcontentsline{0pt}{\bfseries}{app:implementation}
\appcontentsline{1.3em}{\normalfont}{app:components}
\appcontentsline{1.3em}{\normalfont}{app:geometry}
\appcontentsline{1.3em}{\normalfont}{app:motion_state}
\appcontentsline{1.3em}{\normalfont}{app:rollout}
\appcontentsline{1.3em}{\normalfont}{app:controls}
\appcontentsline{0pt}{\bfseries}{app:dataset_usage}
\appcontentsline{0pt}{\bfseries}{app:metrics}
\appcontentsline{1.3em}{\normalfont}{app:metric_settings}
\appcontentsline{1.3em}{\normalfont}{app:reference_metrics}
\appcontentsline{1.3em}{\normalfont}{app:motion_metrics}
\appcontentsline{1.3em}{\normalfont}{app:control_metrics}
\appcontentsline{0pt}{\bfseries}{sec:supp_baselines}
\appcontentsline{1.3em}{\normalfont}{app:motion_conditioning}
\appcontentsline{1.3em}{\normalfont}{app:trajectory_ablations}
\appcontentsline{0pt}{\bfseries}{app:com_vis}
\appcontentsline{1.3em}{\normalfont}{app:perception_evidence}
\appcontentsline{1.3em}{\normalfont}{app:decision_example}
\appcontentsline{0pt}{\bfseries}{sec:limitations}
\endgroup
\clearpage

\section{Implementation Details}
\label{app:implementation}

\subsection{Pretrained Components and Inference Configuration}
\label{app:components}

Kepler4D combines pretrained models for geometry estimation, visual perception,
mask propagation, and video synthesis. Depth Anything 3~\citep{lin2025da3}
provides depth, intrinsics, and camera poses. Qwen2.5-VL-7B-Instruct
\citep{bai2025qwen25vltechnicalreport} is used for scene understanding and
structured motion decisions, while SAM 2.1 Hiera-Large~\citep{ravi2024sam2}
propagates foreground masks. Future video synthesis uses Wan2.1-T2V-14B
\citep{wan2025wan} with the pretrained geometric-conditioning components of
VerseCrafter~\citep{zheng2026versecrafter}. The object trajectories and their
rendered controls connect these components without requiring the video
generator to produce an explicit 3D motion plan itself.

The inference implementation supports the Wan2.1-T2V-14B CFG- and
step-distilled LoRA weights~\citep{hu2022lora} released by
LightX2V~\citep{lightx2v}. The model card attributes their four-step
bidirectional distillation to training code adapted from Self Forcing
\citep{huang2025selfforcing}. The launcher configures a rank-64 v2 LoRA; we do not perform distillation
ourselves.\footnote{\raggedright Upstream model card and LoRA releases:
\href{https://huggingface.co/lightx2v/Wan2.1-T2V-14B-StepDistill-CfgDistill-Lightx2v/tree/main/loras}{\nolinkurl{lightx2v/Wan2.1-T2V-14B-StepDistill-CfgDistill-Lightx2v}, \texttt{loras/}}.
The configured filename is
\href{https://huggingface.co/Kijai/WanVideo_comfy/blob/main/Lightx2v/lightx2v_T2V_14B_cfg_step_distill_v2_lora_rank64_bf16.safetensors}{\nolinkurl{lightx2v_T2V_14B_cfg_step_distill_v2_lora_rank64_bf16.safetensors}},
also published in Kijai's converted LightX2V release.}
Table~\ref{tab:app_inference}
lists the default synthesis configuration exposed by the inference launcher.
These are inference defaults; clip length and the observed/future partition
belong to the evaluation setting rather than to the model architecture. The
reported hardware is an NVIDIA RTX PRO 6000 Blackwell GPU with 96\,GB memory.

\begin{table}[ht]
    \centering
    \caption{Default video-synthesis launcher settings; listed spatial dimensions are height $\times$ width.}
    \label{tab:app_inference}
    \small
    \begin{tabularx}{\linewidth}{@{}p{0.43\linewidth}X@{}}
        \toprule
        Component or setting & Configuration \\
        \midrule
        Video backbone & Wan2.1-T2V-14B \\
        Geometric conditioning & Pretrained VerseCrafter GeoAdapter \\
        Acceleration & LightX2V CFG/step-distillation LoRA; merge weight 1.0 \\
        Numerical precision & bfloat16 \\
        Sampler & Flow-matching Euler scheduler \\
        Denoising steps & 6 \\
        Classifier-free guidance scale & 1.0 \\
        Schedule shift & 5.0 \\
        GeoAdapter context scale & 1.0 \\
        Output resolution & $720\times1280$ \\
        Output frame rate & 16 frames per second \\
        Random seed & 2025 \\
        Negative prompt & None \\
        \bottomrule
    \end{tabularx}
\end{table}

The Qwen motion-decision call uses a temperature of 0.1. Its output is
parsed as structured JSON before trajectory generation. Model loading can be amortized through persistent inference
services; the alternative standalone interface loads the same synthesis
components for each invocation. Consequently, cold-start and warm-service
timings represent different workloads. Runtime depends on the generator,
acceleration settings, and whether model initialization is included.

\subsection{Geometry, Foreground Identification, and Spatial Support}
\label{app:geometry}

\paragraph{Geometric observations.}
Let $C_t^{\mathrm{c2w}}=(R_t,\mathbf{t}_t)$ denote a camera-to-world pose, and let
$\widetilde{\mathbf{u}}=(u_x,u_y,1)^\top$ be a homogeneous pixel coordinate.
The inverse projection used by the state construction can be written as
\begin{equation}
    \mathbf{x}_t(\mathbf{u})
    =R_t\left[D_t(\mathbf{u})K_t^{-1}\widetilde{\mathbf{u}}\right]
    +\mathbf{t}_t.
    \label{eq:app_unprojection}
\end{equation}
This expression makes explicit the roles of depth, calibration, and camera
pose in relating image observations to object motion. Here depth is measured
along the camera's optical axis. Image resizing requires
the corresponding rescaling of focal lengths and principal-point coordinates
before unprojection. Spatial extents and object-center trajectories use the
same reconstructed coordinate system.

For longer observed videos, geometry inference can be performed in overlapping
temporal chunks. The chunking interface uses 60-frame windows with 30-frame
overlap. Adjacent reconstructions are aligned through a similarity transform
estimated from their shared camera poses. The estimated scale is applied to
depth as well as camera translation, so the two remain compatible when
constructing scene points. Related long-sequence reconstruction methods address cross-fragment
consistency through overlap-based alignment~\citep{deng2025vggtlong}
and instance-guided association~\citep{gao2026list3r}; our alignment
is restricted to the observed reconstruction.

\paragraph{Foreground identity and masks.}
The first and last observed frames are provided to the perception VLM. Its
structured output contains object identifiers, semantic labels, endpoint
bounding boxes, a scene description, and scene-context cues. The endpoint
boxes initialize SAM 2 mask propagation across the observed video. Propagation
in both temporal directions allows a prompt on a later frame to contribute to
earlier masks. The resulting object identifiers associate masks across time
and connect each motion decision to the corresponding foreground proxy.
The center average in Eq.~\ref{eq:lifting} is defined for a nonempty mask
with valid depth samples. A fully occluded object has no directly observed
center in that frame.

\paragraph{Coarse spatial support.}
The spatial extent $\mathbf{e}_i$ describes the support used for geometric control, not
a detailed object mesh. A single object keeps the same extent during its
future trajectory, while its center changes. This construction separates
translation guidance from the appearance and local articulation synthesized
by the video generator. In particular, a person's limbs or a vehicle's fine
shape are not explicitly represented by the coarse support.

\subsection{Motion Summaries and Structured Decisions}
\label{app:motion_state}

CoM separates measured motion from semantic decisions. The motion extractor
summarizes displacement, speed, acceleration, and turning trend; the VLM then
chooses how the motion should continue in context. These quantities supplement
the scene description and the endpoint images, rather than requiring the VLM
to recover all numerical motion information directly from pixels.

\paragraph{Displacement and speed.}
For an observed track containing $n_i>1$ equally spaced samples, an endpoint
estimate of its mean velocity is
\begin{equation}
    \mathbf{v}_i
    =\frac{\mathbf{c}_i^{\mathrm{last}}-\mathbf{c}_i^{\mathrm{first}}}{n_i-1},
    \qquad s_i=\|\mathbf{v}_i\|_2.
    \label{eq:app_velocity}
\end{equation}
Motion is expressed in reconstructed coordinate units per sampled frame.
A track with negligible displacement
is assigned zero speed, avoiding division by a nearly zero displacement when
forming its direction. The stationary cue is relative to the object's spatial
extent, so the semantic decision receives both motion magnitude and an
object-scale reference.

\paragraph{Turning and acceleration cues.}
Turning trend is obtained from signed changes between short planar
displacement vectors. The motion extractor uses displacements spanning three
frame intervals in the recent history, rejects nearly zero displacements, takes a
median over valid signed changes, and limits the estimated turn rate to
$[-8,8]$ degrees per sampled frame. The recent-history window contains at most
20 trajectory samples. Acceleration is summarized from changes in recent
step lengths. These estimates are cues for selecting future behavior, rather
than a claim that noisy monocular reconstruction recovers exact physical
velocities or forces.

\paragraph{Decision vocabulary.}
The speed decision $\alpha_i$ selects maintaining, increasing, or decreasing
speed, and the heading decision $\delta_i$ selects straight continuation, a
left turn, or a right turn. As defined in Eq.~\ref{eq:com-decision}, the
response also includes $\eta_i$, which selects whether to apply the lower
bound $h_i$ to the predicted object-center height. This bound constrains the
center trajectory; it does not represent physical contact or enforce an
object mesh to lie on a supporting surface. The motion description $d_i$
provides text guidance for synthesis, associating the geometric controls and
the semantic description with the same predicted behavior.

The JSON parser validates the action vocabulary. An invalid speed action is
replaced by maintaining speed, and an invalid heading action is replaced by
straight continuation. If a tracked object is omitted from a parsed response,
the decision list is completed using these conservative defaults. This
validation addresses the structured control interface; it does not guarantee
that a semantically plausible decision will be inferred in every scene.

\subsection{Speed and Heading Updates in Deterministic Rollout}
\label{app:rollout}

The numerical rollout starts from the latest observed position and estimated
velocity. Here we detail the speed and planar-heading updates within the
rollout interface of the main paper. In Eq.~\ref{eq:traj-rollout}, the
dependence of the motion-parameter mapping on the previously extracted
object-specific motion state $\mathbf{m}_i$ is implicit. The structured
decisions select the motion actions, while the observed turning trend and
semantic cues provide the motion-state information used by the automatic
rollout rules. Let $k=0,\ldots,T_f-1$ index future steps,
and let $g_{\alpha_i}(k)$ be the displacement multiplier associated with the
speed decision. The implemented schedules are
\begin{equation}
    g_{\alpha_i}(k)=
    \begin{cases}
        1, & \text{maintain},\\
        \min(2,1+0.02k), & \text{accelerate},\\
        \max(0.3,1-0.02k), & \text{decelerate}.
    \end{cases}
    \label{eq:app_speed_schedule}
\end{equation}
Here accelerate and decelerate implement the increase and decrease decisions
of the main paper. The multiplier acts on the displacement at each step. It is not repeatedly
multiplied into the stored velocity magnitude, so an acceleration action does
not induce an unintended exponential speed increase.

For a selected per-step turn $\omega_i$, the planar velocity update uses
\begin{equation}
    Q(\omega_i)=
    \begin{bmatrix}
        \cos\omega_i&-\sin\omega_i\\
        \sin\omega_i&\cos\omega_i
    \end{bmatrix},
    \qquad
    \mathbf{v}_{i,\parallel}^{(k+1)}
    =Q(\omega_i)\mathbf{v}_{i,\parallel}^{(k)}.
    \label{eq:app_heading_update}
\end{equation}
In the rollout coordinate convention, $\mathbf{v}_{i,\parallel}$ contains
the $X$ and $Z$ components, and the $Y$ component is unchanged by this rotation.
The initial velocity is $\mathbf{v}_i^{(0)}=\mathbf{v}_i$.
Turn rates stated in degrees are converted to radians before evaluating
$Q(\omega_i)$.
The speed and heading contribution to the position update is then
\begin{equation}
    \widetilde{\mathbf{c}}_i^{(k+1)}
    =\widetilde{\mathbf{c}}_i^{(k)}
    +g_{\alpha_i}(k)\mathbf{v}_i^{(k+1)},
    \qquad
    \widetilde{\mathbf{c}}_i^{(0)}=\mathbf{c}_i^{T_o-1}.
    \label{eq:app_displacement_update}
\end{equation}
Equations~\ref{eq:app_speed_schedule}--\ref{eq:app_displacement_update}
describe the speed and heading substep, with step $k+1$ corresponding to
absolute frame $T_o+k$; the complete motion decision and
trajectory interface are defined in \S~\ref{sec:cop} of the main paper.

For automatic left/right decisions, the measured turn-rate magnitude is used
when it exceeds 0.3 degrees per step; otherwise, a category-derived surface
cue selects a default magnitude. The defaults are 1.5 degrees for water,
2 degrees for road, 3 degrees for ground, and 2 degrees for an unknown cue.
The chosen side determines the sign. Straight continuation uses zero
additional turn. The turn-rate rules above describe automatic continuation.
For language-guided editing, the turning angle specified in the instruction
is passed to the deterministic rollout to construct a corresponding 3D
object-center trajectory. The resulting trajectory is rendered into geometric
controls to guide video synthesis.

The height decision is applied to the object-center coordinate in the local
height frame. When $\eta_i$ enables the constraint, a proposed height below
$h_i$ is raised to $h_i$; otherwise, the proposed height is retained. The
other position coordinates retain their speed-and-heading updates. This is
the optional center-height lower bound in Eq.~\ref{eq:traj-rollout}, not a
projection of the object onto a physical surface. Its behavior depends on
the estimated height reference and the selected decision.

After rollout, a one-dimensional Gaussian filter with a standard deviation
of one sampled frame smooths the
trajectory along time. The first and last generated center positions are
retained from the unsmoothed segment, preserving its endpoints. Smoothing
reduces abrupt variations in the control path; it does not add articulated
motion or simulate forces.

\subsection{Geometric Control Representation}
\label{app:controls}

The renderer represents each foreground object by a Gaussian with mean
$\widehat{\mathbf{c}}_i^t$ and a diagonal covariance derived from its fixed
spatial extent. If $\mathbf{e}_i=(e_{i,1},e_{i,2},e_{i,3})$ contains full axis lengths,
the parameterization is
\begin{equation}
    \Sigma_i=\operatorname{diag}(\sigma_{i,1}^2,\sigma_{i,2}^2,\sigma_{i,3}^2),
    \quad
    \sigma_{i,j}=\max\!\left(\frac{e_{i,j}}{2\rho},10^{-3}\right),
    \qquad \rho=2.5.
    \label{eq:app_covariance}
\end{equation}
When the floor is inactive, the $\rho$-standard-deviation ellipsoid has
half-axis lengths $\mathbf{e}_i/2$. The standard-deviation floor is expressed
in the reconstructed coordinate units and prevents degenerate covariance axes. The covariance
remains fixed during a rollout, while the mean follows the object trajectory.
Changing the center therefore changes the spatial guidance without requiring
an explicit prediction of detailed object shape.

The synthesis interface consumes four control videos and a mask video, as
summarized in Table~\ref{tab:app_controls}. Foreground color identifies the
projected object support; it is not a photorealistic reconstruction of object
texture. Appearance information is supplied by the last observed image and
the pretrained generator. Background and foreground depth provide geometric
context for combining the controls and their visibility.
Here, Gaussians act as coarse trajectory-control proxies, distinct
from Gaussian representations optimized for persistent
tracking~\citep{luiten2024dynamic} or continuous instance
modeling~\citep{wu2026consistent}.

\begin{table}[ht]
    \centering
    \caption{Control streams supplied to the pretrained synthesis interface.
    The combined background/foreground visualization is diagnostic and distinct
    from the separate conditioning streams.}
    \label{tab:app_controls}
    \small
    \begin{tabularx}{\linewidth}{@{}p{0.31\linewidth}X@{}}
        \toprule
        Stream & Information represented \\
        \midrule
        Background RGB & Appearance of the rendered background geometry \\
        Background depth & Depth of the rendered background geometry \\
        Foreground Gaussian RGB & Projected object support and identity color \\
        Foreground Gaussian depth & Depth of the coarse foreground support \\
        Merged mask & Spatial mask used by the video-conditioning interface \\
        \bottomrule
    \end{tabularx}
\end{table}

\paragraph{Appearance and text conditioning.}
The last observed frame $I_{T_o-1}$ is the appearance reference. The synthesis
loader inserts this reference into the first background-conditioning slot
and marks the reference slot accordingly in the input mask. The motion
description produced by CoM supplies the default synthesis prompt. An
explicit prompt or a prompt file can override that description at the
generation interface. Text specifies semantic motion while the geometric
streams specify spatial guidance; neither stream is a substitute for the
other.

\paragraph{Temporal length.}
The video VAE uses a temporal compression factor of four. For a requested
output length $L_{\mathrm{req}}>1$, the synthesis loader uses
\begin{equation}
    L_{\mathrm{out}}
    =4\left\lfloor\frac{L_{\mathrm{req}}-1}{4}\right\rfloor+1.
    \label{eq:app_vae_length}
\end{equation}
Lengths of the form $4k+1$ avoid truncation at this interface. The encoded
clip length, the role of its appearance-reference slot, and the future
prediction horizon are separate bookkeeping quantities. Evaluation compares
the generated continuation with the corresponding held-out portion of the
source video, as described below.

\section{Dataset Usage and Evaluation Settings}
\label{app:dataset_usage}
\label{app:datasets}

\paragraph{Static view synthesis.}
We use the 55-scene DL3DV-Evaluation release of DL3DV~\citep{ling2024dl3dv}
for static evaluation.\footnote{Dataset release and scene list:
\href{https://huggingface.co/datasets/DL3DV/DL3DV-Evaluation}{\texttt{huggingface.co/datasets/DL3DV/DL3DV-Evaluation}}.}
This release is independent of the original DL3DV-10K dataset and distinct
from its 140-scene NVS benchmark. The first 49 frames serve as observations;
held-out views along the provided camera trajectories serve as reconstruction
references.
All methods receive the same observed frames, and methods supporting camera
control receive the same target path. This setting evaluates background
reconstruction and geometry-conditioned synthesis, with no foreground motion
reasoning to assess. Static performance therefore does not independently
measure the accuracy of CoM.

\paragraph{Automatic dynamic continuation.}
The dynamic evaluation draws on the 90 sequences of the DAVIS 2017 train
and validation splits (60 train and 30 validation)
\citep{perazzi2016benchmark,ponttuset2017davis}. Automatic continuation uses
38 of these sequences: 24 train and 14 validation sequences. Each is divided
into an observed prefix and a held-out continuation. The selected sequences
contain 75--104 frames: 37 exceed 75 frames, while \texttt{car-roundabout}
contains 75 frames, with 35 observed and 40 held-out frames. Thus, the
38-sequence set is not defined by a strict greater-than-75-frame rule. All methods receive
the same observed video; no future object trajectory, mask sequence, or
language motion command is supplied as an external control. Descriptions
inferred from the observed video are internal model outputs.

DAVIS does not provide ground-truth camera poses. We estimate camera
trajectories from each full sequence with Depth Anything 3~\citep{lin2025da3} and
provide the future poses as external rendering conditions, shared by all
camera-controllable methods. Thus, this setting evaluates continuation under
a supplied future camera path rather than prediction of the camera motion.
The held-out video provides one realized continuation against which
reference-based metrics are computed.

\paragraph{Language-guided generation.}
The controlled cases are constructed from the same 90-sequence DAVIS 2017
train-and-validation pool. This setting additionally provides a
high-level motion command. Kepler4D converts the command into object-motion
guidance before synthesis.
The text-conditioned baselines receive the same command, without Kepler4D's
inferred trajectories or rendered foreground controls. This comparison
examines the complete route from a shared language instruction to generated
motion, rather than comparing generators supplied with identical geometric
conditioning tensors. Camera-controllable methods use the same DA3-estimated
camera path. Reference-frame comparisons use the corresponding recorded
continuation, while TurnErr provides the restricted camera-turning proxy
defined in Appendix~\ref{app:metrics}.

\paragraph{Temporal correspondence.}
For static evaluation, frame index represents progress along the prescribed
view sequence. Outputs of different lengths are uniformly resampled to the
reference sequence before frame-wise comparison. Dynamic evaluation instead
uses wall-clock time: the common duration of the reference and generated
clips defines the comparison interval, reference-rate timestamps are sampled,
and the nearest generated frame is selected at each timestamp. Selected
frames are resized to the reference resolution. This distinction avoids
equating equal frame indices with equal elapsed time when frame rates differ.

\paragraph{Interpretation of the reference.}
A recorded continuation is useful for measuring similarity, but does not
exhaust the set of plausible futures. In an edited sequence, following a
requested motion may decrease similarity to the unedited recording. We
therefore report reference similarity together with perceptual quality and
motion-related measurements, and interpret each according to the quantity it
actually measures.

\section{Evaluation Metrics}
\label{app:metrics}

\subsection{Metrics by Evaluation Setting}
\label{app:metric_settings}

Static view synthesis uses PSNR, SSIM, LPIPS, and FID to compare held-out
views, together with VBench image quality and temporal-consistency scores.
Automatic dynamic continuation adds Dynamic Degree, Flow-EPE, and RotErr to
characterize motion activity, image motion, and camera rotation. Language
control is evaluated with PSNR, TurnErr, Imaging Quality, and Dynamic Degree.
Trajectory perturbations use AngleErr, SurfErr, Imaging Quality, and Dynamic
Degree; scene-state ablations use Flow-EPE, SurfErr, Dynamic Degree, and
Imaging Quality. These measurements describe complementary properties of the
complete generation pipeline.

\subsection{Reference Similarity and Perceptual Quality}
\label{app:reference_metrics}

\paragraph{PSNR and SSIM.}
PSNR measures pixel-level reconstruction similarity after spatial and temporal
alignment. With RGB values in $[0,255]$, it is computed from the mean squared
error as $10\log_{10}(255^2/\mathrm{MSE})$. SSIM~\citep{wang2004ssim} measures local structural
similarity between corresponding RGB frames. Frame-level measurements are
aggregated within a sequence and then across evaluated sequences. These
metrics compare the generated images with the recorded reference; they do
not establish whether an alternative edited motion is correct.

\paragraph{LPIPS and FID.}
LPIPS~\citep{zhang2018perceptual} measures perceptual distance using the
AlexNet-based implementation, with RGB inputs normalized to $[-1,1]$.
FID~\citep{heusel2017gans} compares the distributions of Inception-V3 features
pooled from aligned generated and reference frames. Lower LPIPS indicates
greater perceptual similarity to the reference, while lower FID indicates a
closer feature distribution. FID is a set-level statistic rather than a
trajectory-following measurement.

\paragraph{VBench.}
We use VBench~\citep{huang2024vbench} in custom-input mode.
Table~\ref{tab:app_vbench} summarizes the dimensions used in the paper.
Imaging Quality and the temporal-consistency measures evaluate properties of
the generated video without requiring its motion to match the recorded
continuation. Dynamic Degree assesses motion activity, so a larger value does
not by itself imply more accurate object motion. The displayed dimensions
are interpreted individually; the Overall column should not be reconstructed
by averaging only the subset of dimensions visible in a results table.

\begin{table}[ht]
    \centering
    \caption{Interpretation of the VBench dimensions used for video evaluation in the main paper.}
    \label{tab:app_vbench}
    \small
    \begin{tabularx}{\linewidth}{@{}p{0.34\linewidth}X@{}}
        \toprule
        Dimension & Measured property \\
        \midrule
        Aesthetic Quality & Visual aesthetic quality \\
        Imaging Quality (Imag.) & Image quality, including sharpness and artifacts \\
        Motion Smoothness (MotSm.) & Smoothness of temporal transitions \\
        Subject Consistency & Consistency of foreground visual identity \\
        Background Consistency (BgC.) & Consistency of background appearance \\
        Dynamic Degree (DynDeg) & Presence of sufficiently strong motion \\
        \bottomrule
    \end{tabularx}
\end{table}

\subsection{Image Motion and Camera Motion}
\label{app:motion_metrics}

\paragraph{Flow-EPE.}
Dense optical flow is estimated with RAFT-Large~\citep{teed2020raft} for both
the generated and reference videos. For $T\geq2$ aligned frames indexed
$0,\ldots,T-1$, let $F_t$ denote flow from frame $t-1$ to frame $t$.
Over image domain $\Omega$, the endpoint error is
\begin{equation}
    \mathrm{Flow\mbox{-}EPE}
    =\frac{1}{T-1}\sum_{t=1}^{T-1}
    \frac{1}{|\Omega|}\sum_{\mathbf{u}\in\Omega}
    \left\|F_t^{\mathrm{gen}}(\mathbf{u})-
    F_t^{\mathrm{ref}}(\mathbf{u})\right\|_2.
    \label{eq:app_flow}
\end{equation}
The flow field is evaluated over the image, so this quantity reflects the
combined image motion caused by camera movement, object displacement, and
local appearance changes. It does not isolate foreground trajectory
prediction from the rest of the generation pipeline.

\paragraph{RotErr.}
We adapt the rotation error of CameraCtrl~\citep{he2024cameractrl}, which
sums per-frame geodesic errors over a clip. Our version averages these errors
and reports them in degrees, avoiding a direct dependence on the number of
evaluated frames. We compare camera rotations estimated by DA3-Large from
generated and reference videos. For aligned frames indexed $0,\ldots,T-1$, $R_t$ below
denotes the camera rotation relative to the first frame, so
$R_0^{\mathrm{gen}}=R_0^{\mathrm{ref}}=I$.
The rotation error is the mean geodesic distance over the remaining frames:
\begin{equation}
    \mathrm{RotErr}
    =\frac{180}{\pi(T-1)}\sum_{t=1}^{T-1}
    \arccos\!\left(
    \operatorname{clip}\!\left[
    \frac{\operatorname{tr}(R_t^{\mathrm{gen}}
    (R_t^{\mathrm{ref}})^\top)-1}{2},-1,1\right]\right).
    \label{eq:app_rotation}
\end{equation}
The factor $180/\pi$ expresses the result in degrees. Clipping the arccosine
argument prevents numerical overflow outside its valid interval. This metric
measures camera rotation rather than the heading of an independently moving
foreground object.

\subsection{Control-Related Measurements}
\label{app:control_metrics}

\paragraph{TurnErr.}
The turning proxy compares a commanded angle $\theta_{\mathrm{cmd}}$ with a
camera-yaw quantity estimated from the generated clip:
\begin{equation}
    \mathrm{TurnErr}
    =\left|\widehat{\theta}_{\mathrm{gen}}-\theta_{\mathrm{cmd}}\right|,
    \qquad
    \widehat{\theta}_{\mathrm{gen}}
    =\max_t\left|\sum_{\tau=1}^{t}\Delta\widehat{\psi}_\tau\right|.
    \label{eq:app_turn}
\end{equation}
Here $\Delta\widehat{\psi}_\tau$ is the estimated camera-yaw increment;
all angles are in degrees and $\theta_{\mathrm{cmd}}\geq0$ denotes the
commanded turning magnitude.
For the car-roundabout example, background features are tracked across
consecutive frames and a RANSAC homography rejects outliers. Median horizontal
feature displacement is converted to a per-frame yaw change and accumulated
over the clip. This yields the camera-motion signal used by the proxy. As specified in the main paper, TurnErr reflects
camera turning and does not independently recover object heading. Its use of
an absolute peak also means that it does not, on its own, distinguish left
from right or a final angle from a transient maximum. TurnErr is reported as
a diagnostic of generated camera turning and is not used to establish
object-level compliance with the motion instruction. In particular, when the
target camera path is fixed, accurate object turning does not necessarily
imply a low TurnErr. Figure~\ref{fig:control_compare} provides qualitative
examples of the object-motion responses to different instructions.

\paragraph{AngleErr.}
In the trajectory analysis, AngleErr compares the specified motion direction
$\mathbf{d}_{\mathrm{ctrl}}$ with the direction
$\mathbf{d}_{\mathrm{gen}}$ obtained from reconstructed, mask-tracked generated
object centers:
\begin{equation}
    \mathrm{AngleErr}
    =\arccos\!\left(
    \operatorname{clip}\!\left[
    \frac{\mathbf{d}_{\mathrm{gen}}^\top\mathbf{d}_{\mathrm{ctrl}}}
    {\|\mathbf{d}_{\mathrm{gen}}\|_2\|\mathbf{d}_{\mathrm{ctrl}}\|_2},-1,1
    \right]\right).
    \label{eq:app_angle}
\end{equation}
Angles are converted to degrees for reporting. This angular comparison
is defined for nonzero directions expressed in a common coordinate frame
and concerns adherence to a supplied control direction,
rather than the ability to predict the reference future trajectory from
observations.

\paragraph{SurfErr.}
Using the notation of the evaluation, $z_t^{\mathrm{gen}}$ is the generated
object-center height in the local support coordinate and $h_{\mathrm{sup}}$
is the estimated object-center support height inferred from the observed
trajectory. The reported quantity is
\begin{equation}
    \mathrm{SurfErr}
    =\frac{1}{T}\sum_{t=1}^{T}
    \left|z_t^{\mathrm{gen}}-h_{\mathrm{sup}}\right|.
    \label{eq:app_surface}
\end{equation}
Here $z_t^{\mathrm{gen}}$ denotes the scalar height, not the raw $Z$ component
used in the rollout plane. The quantity has units of the reconstructed height
coordinate and measures deviation from an estimated center-height reference. It does
not compare the object's boundary with a physical supporting surface and
does not measure contact accuracy. Lower SurfErr indicates closer adherence
to this estimated height reference. Its interpretation depends on the
reconstruction, the tracked object centers, and the chosen local coordinate.
Here $h_{\mathrm{sup}}$ denotes the evaluation reference, while $h_i$ in
\S~\ref{sec:cop} is the optional lower-bound parameter of trajectory rollout.

\section{Baseline Scope and Ablation Interpretation}
\label{sec:supp_baselines}
\label{app:baselines}

The baselines test complementary ways of connecting observations, geometry,
and video generation. We use the same observed video and, when supported,
the same target camera trajectory for all methods. We follow official
inference pipelines where available, adapting input formatting, frame
resolution, and camera conventions to the evaluation setting. SPMem~\citep{wu2025video} provides spatial memory for
scene-consistent generation.
This generative use of memory differs from static-scene storage
and retrieval across revisits~\citep{wang2026cognimap3d}.
GEN3C~\citep{ren2025gen3c} uses a 3D cache as
geometric guidance. TrajectoryCrafter~\citep{yu2025trajectorycrafter} redirects
the camera trajectory of a monocular source video. VerseCrafter
\citep{zheng2026versecrafter} provides joint camera/object geometric controls,
including object trajectories. These capabilities differ from inferring a
future object trajectory through CoM and then rendering that trajectory for
synthesis.

The language-control comparison also includes Wan Inpaint, which uses the
Wan2.1 family~\citep{wan2025wan}. The label denotes the baseline's use in this
experiment rather than a separate model family. In the language-guided
setting, its motion instruction is supplied as text. It does not receive the
object trajectories or foreground control maps inferred by Kepler4D.

\subsection{Motion Description and Trajectory Guidance}
\label{app:motion_conditioning}

The motion-conditioning study separates a description of the observed scene
from a description of future motion, and then adds spatial trajectory
guidance. Table~\ref{tab:app_motion_ablation} summarizes this distinction.
The comparison asks what the generation pipeline gains from these different
forms of motion information; it is not a direct evaluation of a VLM's
standalone 3D forecasting accuracy.

\begin{table}[ht]
    \centering
    \caption{Text and trajectory guidance supplied by the three tested motion-conditioning variants.}
    \label{tab:app_motion_ablation}
    \small
    \begin{tabularx}{\linewidth}{@{}p{0.25\linewidth}XX@{}}
        \toprule
        Variant & Text guidance & CoM trajectory guidance \\
        \midrule
        Qwen prompt & Observed scene description & Absent \\
        CoM w/o traj. & Predicted motion description & Not rendered into foreground controls \\
        CoM w/ traj. & Predicted motion description & Rendered future object trajectories \\
        \bottomrule
    \end{tabularx}
\end{table}

\subsection{Trajectory Perturbations and Scene-State Components}
\label{app:trajectory_ablations}

The trajectory-control analysis uses a reference trajectory derived from the
source video and directional perturbations of that trajectory. It examines
the response of the synthesis pipeline to changed geometric guidance.
AngleErr and deviation from the estimated object-center support height characterize properties
of the generated response. This study is distinct from automatic motion
prediction, because a source-derived trajectory is supplied as a control.

The scene-state component study examines observed motion history, semantic
context, and surface-related cues within the same generation pipeline.
Motion history describes how the object has moved; semantic context describes
its identity and surroundings; surface-related cues provide additional
context for interpreting likely motion. The reported changes in Flow-EPE,
DynDeg, and Imaging Quality are interpreted as effects on the complete
pipeline. In particular, a change in Dynamic Degree measures a change in
motion activity, and does not by itself establish that one trajectory
prediction is more accurate.

\input{evolveworld_nips/figs/tex/com.tex}

\section{Visualization of Chain-of-Motion Trajectory Prediction}
\label{app:com_vis}
\label{app:interface}

Figure~\ref{fig:com_vis} schematically illustrates the scene information,
motion state, and structured decisions used by Chain-of-Motion to guide
deterministic future 3D object-center trajectory rollout. The scene state records
background geometry, dynamic-object identities, observed object centers and
extents, and semantic scene context. These quantities connect the input
video to a structured motion interface in the reconstructed 3D frame.

The \texttt{surface\_adherence} field in the schematic example denotes the
decision $\eta_i$ in Eq.~\ref{eq:com-decision}. In our rollout, it enables
the estimated object-center height lower bound $h_i$, rather than enforcing
contact with a reconstructed road surface.

For each object, motion-state extraction summarizes the latest center,
velocity, speed, acceleration, turning trend, height threshold, coarse
extent, and semantic cues, as in Eq.~\ref{eq:motion-state}. CoM then predicts
speed and heading decisions, the optional center-height lower-bound decision,
and a textual motion description. Deterministic rollout converts the
structured decisions into future center positions. The predicted or edited
trajectory becomes rendered foreground guidance under the target camera
path, while the motion description provides semantic guidance for synthesis.
The generator supplies appearance and local articulation beyond this coarse
trajectory representation.

The interface below provides schematic examples of the connection shown in
Figure~\ref{fig:com_vis}. These examples describe fields and their roles;
they are not numerical logs from a benchmark sequence.

\subsection{Perception and Motion Evidence}
\label{app:perception_evidence}

The perception query asks the VLM to identify foreground objects in the first
and last observed frames, assign the same identity across the two frames,
return endpoint boxes, and summarize the scene. Object boxes then initialize
mask tracking. The motion query augments the resulting description with
measured direction, speed, acceleration, turning trend, and stationary cues.
Table~\ref{tab:app_interface} relates these quantities to their roles.

\begin{table}[ht]
    \centering
    \caption{Roles of observed information and motion decisions in the structured CoM interface.}
    \label{tab:app_interface}
    \small
    \begin{tabularx}{\linewidth}{@{}p{0.33\linewidth}X@{}}
        \toprule
        Information & Role \\
        \midrule
        Endpoint images & Visual evidence for object identity and scene context \\
        Object identifier and label & Associate masks, history, decisions, and controls \\
        Observed motion summary & Numerical evidence for speed and heading decisions \\
        Scene description & Context for a plausible continuation \\
        Speed, heading, and height-bound decisions & Inputs to deterministic trajectory updates \\
        Motion description & Semantic guidance for video synthesis \\
        \bottomrule
    \end{tabularx}
\end{table}

\subsection{Example of a Motion Decision}
\label{app:decision_example}

Consider a tracked car whose observed history indicates forward movement
along a road. A schematic motion query can be expressed as follows:

\begin{quote}
    Given the observed scene, the endpoint images, and the measured motion of
    each tracked object, select speed and heading actions and decide whether
    to apply the estimated lower height bound to its future center positions.
    Return the decisions with the corresponding object identifier and a concise
    description of the expected motion for video synthesis.
\end{quote}

A possible motion-related subset of the response is

\begin{center}
    \begin{tabular}{@{}ll@{}}
        \toprule
        Field & Schematic value \\
        \midrule
        Object identifier & 1 \\
        Object label & Car \\
        Speed action & Maintain \\
        Heading action & Straight \\
        Height-bound decision & Enabled \\
        Motion description & The car continues moving forward along the road. \\
        \bottomrule
    \end{tabular}
\end{center}

The deterministic rollout turns the selected actions and the observed motion
state into future center positions, applying the height threshold only when
the corresponding decision enables it. Those positions become geometric
foreground guidance, while the motion description becomes text conditioning.
The two outputs serve different purposes: the trajectory specifies where the
coarse object support moves, and the description conveys the intended
behavior to the pretrained generator. Fine appearance and articulation are
synthesized during video generation.

\begingroup
\setlength{\parskip}{0pt}
\section{Limitations and Future Work}
\label{sec:limitations}
\label{app:limitations}

\paragraph{Scope of the motion representation.}
Kepler4D targets editable object-center motion under a supplied camera path.
This compact interface lets the same trajectory support motion prediction,
language editing, and geometric rendering, while the pretrained generator
supplies appearance and local articulation. Fixed coarse supports do not
explicitly model joint motion, shape deformation, or contact dynamics, so
geometric guidance alone does not guarantee physically correct interactions.
Extending the interface with articulated supports and contact constraints is
a natural direction for applications that require finer physical control.

\paragraph{Reliability of reconstructed state.}
The scene state is estimated from monocular video using pretrained depth,
camera, segmentation, and language models. Occlusion and changes in visible
surface area can shift a reconstructed center even when the physical center
moves differently. Kepler4D exposes these estimates as explicit motion
histories, making their role in subsequent decisions inspectable, but does
not currently propagate calibrated uncertainty through the rollout.
Confidence-aware tracking and joint refinement of geometry and motion offer
ways to improve robustness to difficult observations without changing the
trajectory-based control interface.
Related directions include geometry estimation in dynamic
scenes~\citep{zhang2025monst3r} and motion-aware streaming
reconstruction~\citep{wang2026raymap3r}; their integration and
downstream effects in Kepler4D remain to be evaluated.

\endgroup

%% file: evolveworld_nips/figs/tex/com.tex
\begin{figure}[t]
    \centering
    \includegraphics[width=1.0\textwidth]{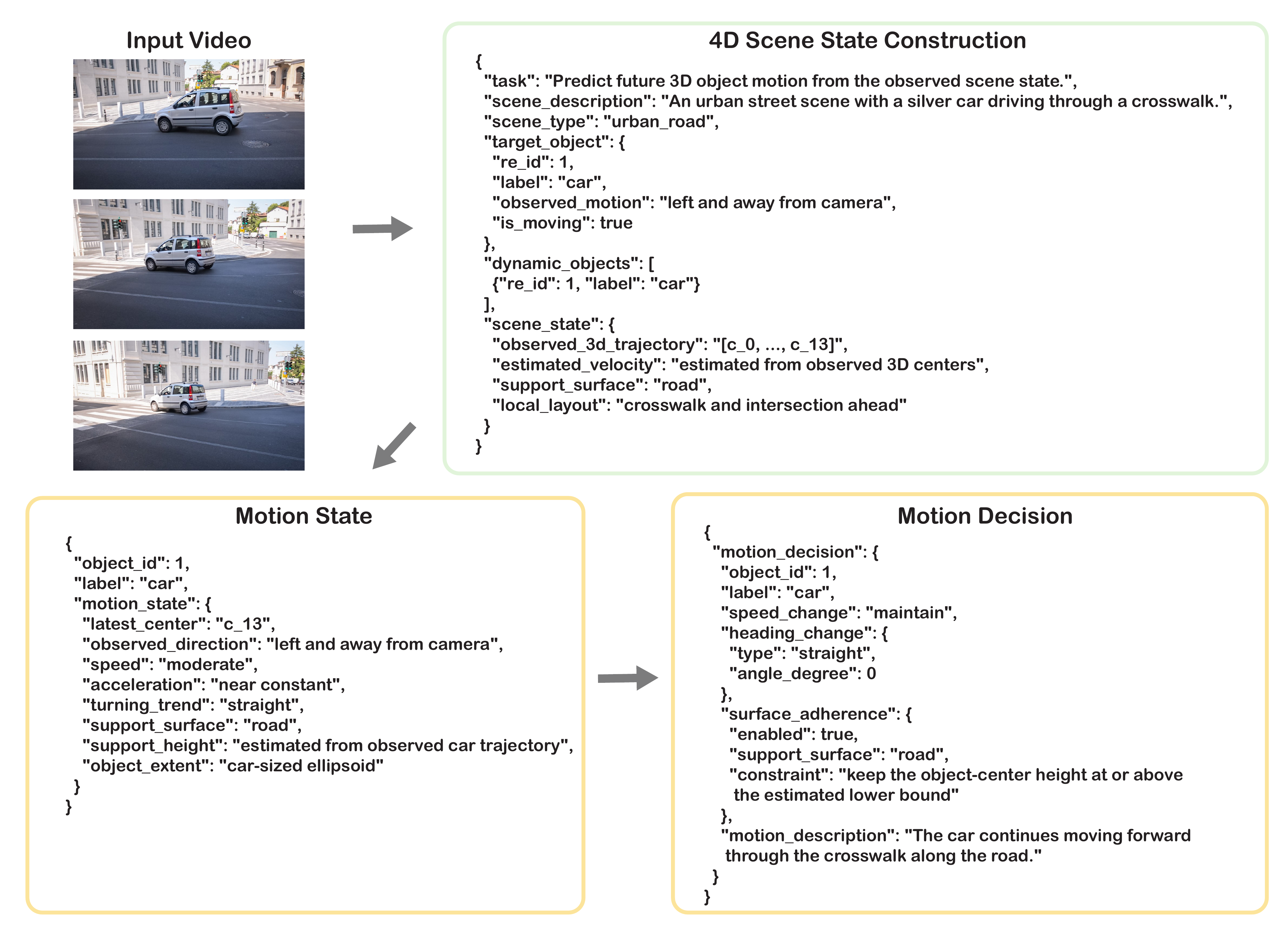}
    \caption{\textbf{Schematic illustration of Chain-of-Motion inputs and outputs.}
    CoM uses observed motion and semantic scene context to select speed and heading actions, decide whether to apply a lower bound to object-center height, and produce a motion description.
    The structured decisions guide deterministic 3D trajectory rollout, while the description provides text conditioning for video synthesis.
    The fields illustrate the interface and are not numerical logs from a benchmark sequence.}
    \label{fig:com_vis}
    \vspace{-0.5em}
\end{figure}